\documentclass[10pt,journal,cspaper,compsoc]{IEEEtran}
\ifCLASSOPTIONcompsoc
   \usepackage[nocompress]{cite}
\else
\fi
\usepackage{epsfig}
\usepackage{graphicx}
\usepackage{amsmath}
\usepackage{amssymb}
\usepackage{epsfig}
\usepackage{graphicx}
\usepackage{amsmath}
\usepackage{amssymb}
\usepackage{epstopdf}
\usepackage{algorithm}
\usepackage{algorithmic}
\usepackage{subfigure}

\begin{document}
%
\title{Fast Iteratively Reweighted Least Squares Algorithms for Analysis-Based Sparsity Reconstruction}
%
%
%
%

\author{Chen~Chen,~\IEEEmembership{Student Member,~IEEE,}
        ~Junzhou~Huang*,~\IEEEmembership{Member,~IEEE,}
        ~Lei~He, 
        and~Hongsheng~Li
\IEEEcompsocitemizethanks{\IEEEcompsocthanksitem C. Chen and J. Huang are with the Department of Computer Science and Engineering, University of Texas at Arlington, Texas 76019, USA.
Corresponding author: Junzhou Huang. Email: jzhuang@uta.edu.
\IEEEcompsocthanksitem L. He is with the Office of Strategic Initiatives, Library of Congress, 101 Independence Ave. SE, Washington, DC 20540.
\IEEEcompsocthanksitem H. Li is with the School of Electronic Engineering at University of Electronic Science and Technology of China, 2006 Xi-Yuan Ave., Chengdu, China.}
\thanks{}}

%
%

\markboth{ IEEE TRANSACTIONS ON PATTERN ANALYSIS AND MACHINE INTELLIGENCE, ,~Vol.~XX, No.~XX, XX~XXXX}%
{Chen \MakeLowercase{\textit{et al.}}: FIRLS for analysis-based sparsity reconstruction}
%


\IEEEcompsoctitleabstractindextext{%
\begin{abstract}
In this paper, we propose a novel algorithm for analysis-based sparsity reconstruction. It can solve the generalized problem by structured sparsity regularization with an orthogonal basis and total variation regularization. The proposed algorithm is based on the iterative reweighted least squares (IRLS) model, which is further accelerated by the preconditioned conjugate gradient method.  The convergence rate of the proposed algorithm is almost the same as that of the traditional IRLS algorithms, that is, exponentially fast. Moreover, with the specifically devised preconditioner, the computational cost for each iteration is significantly less than that of traditional IRLS algorithms, which enables our approach to handle large scale problems. In addition to the fast convergence, it is straightforward to apply our method to standard sparsity, group sparsity, overlapping group sparsity and TV based problems. Experiments are conducted on a practical application: compressive sensing magnetic resonance imaging. Extensive results demonstrate that the proposed algorithm achieves superior performance over 14 state-of-the-art algorithms in terms of both accuracy and computational cost.
\end{abstract}

}

\maketitle

\IEEEdisplaynotcompsoctitleabstractindextext

%
\IEEEpeerreviewmaketitle

\section{Introduction}

\IEEEPARstart{I}ll-posed problems widely exist in medical imaging and computer vision. In order to seek a meaningful solution, regularization is often used if we have certain prior knowledge.
With the emerging of compressive sensing (CS) \cite{Candes06}, sparsity regularization has been an active topic in recent years.
If the original data is sparse or compressible, it can be recovered precisely from a small number of measurements.
The $\ell_1$ norm is usually used to induce sparsity and gains great success in many
real applications.
The optimization problems can be written as:
\begin{eqnarray}
\min_x \{F(x) = \frac{1}{2}||Ax-b||_2^2+\lambda ||x||_1\},
 \label{eqn:l1pro}
\end{eqnarray}
where $A \in \mathbb{R}^{M \times N}$ is the measurement matrix and $b\in \mathbb{R}^{M}$ is the vector of measurements; $x\in \mathbb{R}^{N}$ is
the data to be recovered; $\lambda$ is a positive parameter.

According to structured sparsity theories \cite{Baraniu10,Huang11}, more benefits can be achieved if we could utilize more prior information about the sparsity patterns. For example, the components of the data may be clustered in groups, which is called group sparse data.
Components within the same group tend to be zeros or
non-zeros. Sometimes one component may appear in several
groups simultaneously, which corresponds to the overlapping
group sparsity \cite{jacob2009group}. A favorable method would be replacing the
$\ell_1$ norm with $\ell_{2,1}$ norm to model the group sparsity \cite{yuan2005model}:
\begin{eqnarray}
 ||x||_{2,1} = \sum ||x_{g_i}||_2, \quad i=1,2,...,s,
 \label{eqn:l21}
\end{eqnarray}
where $x_{g_i}$ denotes the components in $i$-th group and $s$ is
the total number of groups.  It has been proven that, less measurements are required for structured sparsity recovery, or more precise solution can be obtained with the same number of measurements \cite{Baraniu10,Huang11,bach2011optimization}.

In many real-world applications, the data itself is not sparse, but it can be sparsely represented in some transformation domains. This leads to the analysis-based sparsity regularization problem:
\begin{eqnarray}
\min_x \{F(x) = \frac{1}{2}||Ax-b||_2^2+\lambda ||\Psi x||_{2, 1}\},
 \label{eqn:anapro}
\end{eqnarray}
where $\Psi$ denotes some sparifying basis, e.g. the wavelet or finite difference basis. In this article, we are interested in this generalized sparsity regularization problem (\ref{eqn:anapro}), which may contain overlapped groups.
The standard sparsity and non-overlapping group sparsity minimization problem are special cases of problem (\ref{eqn:anapro}). In our work, we focus on the image reconstruction applications, e.g. CS imaging \cite{xiao2010fast}, image inpainting \cite{bertalmio2003simultaneous}, compressive sensing magnetic resonance imaging (CS-MRI) \cite{lustig2007sparse}, where $A$ is an undersampling matrix/operator.


When $\Psi$ is an orthogonal basis, problem (\ref{eqn:anapro}) corresponds to the Lasso problem \cite{tibshirani1996regression}.
In the literature, many efficient algorithms can be used to solve the standard Lasso and non-overlapping group Lasso, such as FISTA \cite{Beck09}, SPGL1
\cite{Bergi08JSC}, SpaRSA \cite{Wright09}, FOCUSS \cite{FOCUSS}. However, there are relatively much fewer algorithms for overlapping group Lasso, due to the non-smoothness and non-separableness of the overlapped $\ell_{2,1}$ penalty. SLEP \cite{SLEP,yuan2013efficient}, GLO-pridu \cite{mosci2010primal} solve the overlapping group sparsity problem by identifying active groups, and YALL1 \cite{Deng11} solves it with the alternating direction method (ADM). Both SLEP and GLO-pridu are based on the proximal gradient descent method (e.g. FISTA \cite{Beck09}), which cannot achieve a convergence rate better than $F(x^k)-F(x^*)\sim\mathcal{O}(1/k^2)$, where $x^*$ denotes the optimal solution and $k$ is the iteration number. YALL1 relaxes
the original problem with augmented Lagrangian and iteratively
minimizes the subproblems based on the variable splitting method. Generally, the
convergence rate of ADM is no better than $\mathcal{O}(1/k)$ in sparse recovery
problems. Although they are very efficient in each iteration, a large number of iterations may be required due to the relatively slow convergence rate.
On the other hand, the iterative reweighted least squares (IRLS)
algorithms have been proven that they converge exponentially fast \cite{Daubechies08} \cite{Chartrand08}. Unfortunately, conventional IRLS algorithms contain a large scale inverse operation in each step, which makes them still much more computationally expensive than the fastest proximal methods such as FISTA \cite{bach2011optimization}. In addition, it is unknown how to extend these IRLS based algorithms to solve the overlapping group Lasso problems. Some other algorithms can solve the sparsity or group sparsity based denoising problems efficiently \cite{chen2014translation,chen2014group}, but they 
can not solve the general linear inverse problem (\ref{eqn:anapro}) directly.

Another special case of (\ref{eqn:anapro}) is the total variation (TV) reconstruction problem, where $\Psi$ denotes the first-order finite difference matrices and is non-orthogonal.
There are several efficient algorithms specially designed for TV reconstruction, including the RecPF \cite{yang2010fast} and SALSA \cite{afonso2010fast}. Both of them are relaxed by the ADM. The efficient transformation in RecPF \cite{yang2010fast} requires that $A^TA$ can be diagonalized by the Fourier transform, while SALSA \cite{afonso2010fast} requires $AA^T = I$.
Due to these restrictions, these two methods can not be applied to certain reconstruction applications, e.g. CS imaging \cite{xiao2010fast}.
Moreover, it is unknown how to extend them to solve the joint total variation (JTV) problems. 
Moreover, the ADM-based methods often have slower convergence rate.
Of course, generalized minimization methods can be used, such as the split Bregman method \cite{goldstein2009split}, FISTA \cite{beck2009fast} and IRN \cite{rodriguez2009efficient}, but they have their own inferiority without considering the special structure of undersampling matrix $A$ in reconstruction.

In this article, we propose a novel scheme for the analysis-based sparsity reconstruction (\ref{eqn:anapro})
based on the IRLS framework. 
It preserves the fast convergence performance of traditional IRLS, which only requires
a few reweighted iterations to achieve an accurate solution. We call our method fast iterative reweighted least squares (FIRLS).
Moreover, we propose a new ``pseudo-diagonal" type preconditioner to significantly accelerate the inverse
subproblem with preconditioned conjugate gradient (PCG) method. This preconditioner is based on the observation that $A^TA$ is often diagonally dominant in the image reconstruction problems. With the same computation complexity, the proposed preconditioner provides more precise results than conventional Jacobi diagonal preconditioner. In addition,
the proposed preconditioner can be applied even when $A$ is an operator, e.g., the Fourier or wavelet transform, which is not feasible for most existing preconditioners of the PCG methods.
Besides the efficiency and fast convergence rate, the proposed method can
be easily applied to different sparsity patterns,  e.g. overlapping group sparsity, TV. 
We validate the proposed method on CS-MRI for tree sparsity, joint sparsity, and TV based reconstruction.
Extensive experimental results demonstrate that the proposed algorithm outperforms the state-of-the-art methods in terms of both accuracy and computational speed. Part of results in this work has been presented in \cite{chen2014preconditioning}.

\section{Related Work: IRLS}


The conventional IRLS algorithms solve the standard sparse problem in this constrained form:
\begin{eqnarray}
\min_x ||x||_1, \textup{subject } \textup{to }Ax = b.
 \label{eqn:uncL1}
\end{eqnarray}
In practice, the $\ell_1$ norm is replaced by a reweighted $\ell_2$ norm \cite{Chartrand08}:
\begin{eqnarray}
\min_x x^TWx, \textup{subject } \textup{to }Ax = b.
 \label{eqn:IRLS}
\end{eqnarray}
The diagonal weight matrix $W$ in the $k$-th iteration is computed from the solution of the current iteration $x^{k}$, in particular, the diagonal elements $W^k_i = |x_i^{k}|^{-1}$. With current weights $W^k$, we can derive the closed form solution for $x^{k+1}$:
\begin{eqnarray}
x^{k+1} = (W^k)^{-1}A^T(A(W^k)^{-1}A^T)^{-1}b.
 \label{eqn:xn}
\end{eqnarray}
The algorithm can be summarized in Algorithm \ref{alg:IRLS}. It has been proven that the IRLS algorithm converges exponentially fast under mild conditions \cite{Daubechies08}:
\begin{eqnarray}
||x^k-x^*||_1 \leq \mu ||x^{k-1}-x^*||_1\leq \mu^k ||x^{0}-x^*||_1,   \label{eqn:linc}
\end{eqnarray}
where $\mu$ is a fixed constant with $\mu<1$. 
However, this algorithm is rarely used in compressive sensing
applications especially for large scale problems.
That is because the inverse of $A(W^k)^{-1}A^T$ takes $\mathcal{O}(M^3)$ if $A$ is a $M\times N$ sampling matrix. Even with higher convergence rate, traditional IRLS still cannot compete with the
fastest first-order algorithms such as FISTA \cite{Beck09} (some results have been shown in \cite{bach2011optimization}). Moreover, none of existing IRLS methods \cite{Chartrand08,Daubechies08,FOCUSS} could solve the overlapping group sparsity problems, which significantly limits the usage.

 \begin{algorithm}[H]
\caption{IRLS} \label{alg:IRLS}
\begin{algorithmic}
   \STATE {\bfseries Input:} $A$,$b$,$x^1$,$k=1$
   \WHILE {not meet the stopping criterion}
   \STATE Update $W$: $W^k_i = |x_i^{k}|^{-1} \quad \forall \ W^k_i $
   \STATE Update $x$: $x^{k+1} = (W^k)^{-1}A^T(A(W^k)^{-1}A^T)^{-1}b$
   \STATE Update $k=k+1$
   \ENDWHILE
\end{algorithmic}
\end{algorithm}

%
%
%

\section{FIRLS for overlapping group sparsity}

\subsection{An alternative formulation for overlapping group sparsity}

\begin{figure}[htbp]
\centering \vspace{-0.0cm}
 \includegraphics[scale=0.55]{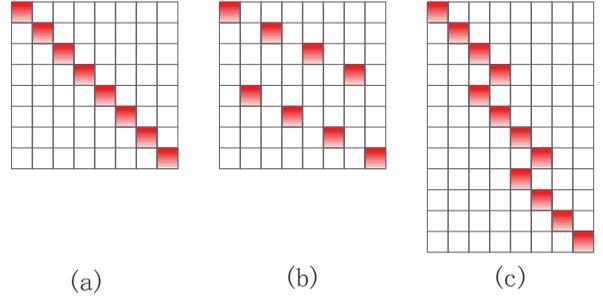}
\caption{Examples of group configuration matrix $G$ for a signal of size
8. The red elements denote ones and white elements denote zeros. (a)
standard sparsity case where $G$ is the identical matrix. (b)
non-overlapping groups of [1,3,5,7] and [2,4,6,8]. (c) overlapping
groups of [1,2,3,4], [3,4,5,6] and [5,6,7,8]. Their group sizes are
$1$,$4$ and $4$, respectively. }\label{fig:G}\vspace{-0.0cm}
\end{figure}

We consider the overlapping group Lasso problem first \cite{yuan2005model,jacob2009group}. 
The mixed $\ell_{2,1}$ norm in (\ref{eqn:anapro}) may contain overlapping groups. It can be rewritten in the analysis-based sparsity form:
\begin{eqnarray}
\min_x \{F(x) = \frac{1}{2}||Ax-b||_2^2+\lambda || G \Phi x||_{2,1}\},
 \label{eqn:probogs}
\end{eqnarray}
where $\Phi$ denotes an orthogonal sparse basis and is optional. A good choice of $\Phi$ for natural images/signals
would be the orthogonal wavelet transform. $G$ is a binary matrix for group configuration, which is constructed by the rows of the identity matrix.
With different settings of $G$, this model can handle overlapping
group, non-overlaping group and standard sparsity problems. Tree sparsity can also be approximated by this model \cite{kim2010tree,liu2010moreau,jenatton2011proximal}.
Simple examples of $G$ for different types of group sparse problems are
shown in Fig. \ref{fig:G}. Although $G$ may have large
scales, it can be efficiently implemented by a sparse matrix. This kind of indexing matrix has been used in the previous
work YALL1 \cite{Deng11}. With this reformulation, $\Psi = G \Phi$ and the $\ell_{2,1}$ norm in (\ref{eqn:probogs}) is now non-overlapping.

Consider the Young's inequality holding for a general function $g(\cdot):\mathbb{R}\longmapsto\mathbb{R}$:
\begin{eqnarray}
\sqrt{g(x)}\leq \frac{\sqrt{g(y)}}{2} +\frac{g(x)}{2\sqrt{g(y)}},
 \label{eqn:gx}
\end{eqnarray}
with $g(y) >0$ and $g(x)\geq 0$. The equality holds only when $g(x)=g(y)$. Based on this, we have
\begin{align}
||G\Phi x||_{2,1} =& \sum_{i=1}^s ||(G\Phi x)_{g_i}||_2  \notag \\
\leq \sum_{i=1}^s &\big[\frac{||(G\Phi x^k)_{g_i}||_{2}}{2} + \frac{||(G\Phi x)_{g_i}||_2^2}{2||(G\Phi x^k)_{g_i}||_2}\big].
 \label{eqn:g21}
\end{align}
Writing it in matrix form and we can majorize $F(x)$ by the majorization minimization (MM) method \cite{Hunter04}:
\begin{eqnarray}
Q(x,W^{k}) = \frac{1}{2}||Ax-b||_2^2+\frac{\lambda}{2}x^T \Phi^T G^TW^kG\Phi x
 \notag \\
+ \frac{\lambda}{2} \sum_{i=1}^s \frac{1}{W^k_{g_i}},
 \label{eqn:GWx}
\end{eqnarray}
where $\Phi^T$ denotes the inverse transform of $\Phi$; $W^{k}$ is the group-wise weights. The weight of $i$-th group $W^k_{g_i}$ can be obtained by:
\begin{eqnarray}
W^k_{g_i} = (||(G \Phi x^{k})_{g_i}||^2_{2}+\epsilon)^{-1/2}.
\label{eqn:gw}
\end{eqnarray}
$\epsilon$ is very small number (e.g. $10^{-10}$) to avoid infinity. Suppose that the signal $x$ to be
recovered is of length $N$ and $G$ is a $N'$-by-$N$ matrix, then
$W^{k}$ is a $N'$-by-$N'$ diagonal matrix and has the following form:
\begin{eqnarray}
W^{k} =\begin{Bmatrix}
W^k_{g_1}  &  &  &  & & \\
 &  ... &  &  & &\\
  &  & W^k_{g_1}  &  & &\\
 &  &  & ... & &\\
 &  &  & &   W^k_{g_s} & \\
 &  &  & & & W^k_{g_s}
\end{Bmatrix},
\label{eqn:Wog}
\end{eqnarray}
where each group-wise weight $W^k_{g_i}$ is duplicated $|g_i|$ times and $|g_i|$ denotes the size of the $i$-th group.
One can find that the group-wise weights are all related to $G$.
With different settings of $G$, the group-wise weights are
directly derived. Variant-size group sparsity problems also can be flexibly
handled in this model. An interesting case would be the standard
sparse problem, where each group contains only one element and
the group-wise weight matrix $W$ is the same as that in IRLS
algorithm \cite{Daubechies08,Chartrand08}.

Now the problem becomes:
\begin{eqnarray}
x^{k+1} = \arg\min_x Q(x,W^{k}).
 \label{eqn:lostf}
\end{eqnarray}
Note that $W^k_{g_i}$ is independent of $x$ and can be considered as a
constant. We iteratively update $W^k$ with $x^{k}$ and solve $x^{k+1}$
based on current $W^k$. Our algorithm is also a IRLS type algorithm with
exponentially fast convergence rate.

\subsection{Accelerating with PCG}
In each iteration, $W^k$ can be easily updated with (\ref{eqn:Wog}) and
(\ref{eqn:gw}). To solve (\ref{eqn:lostf}), a simple way is to let
the first order derivative of $Q(x|x^{k})$ be zero as it is a quadratic
convex function:
\begin{eqnarray}
(A^TA+\lambda\Phi^TG^TW^kG\Phi)x^{} - A^Tb = 0.
 \label{eqn:subprob}
\end{eqnarray}
The way to solve (\ref{eqn:subprob}) determines the efficiency of the
whole algorithm. The exact inverse of the system matrix $S =
A^TA+\lambda\Phi^TG^TW^k G \Phi$ takes $\mathcal{O}(N^3)$
time. It is impractical to compute $S^{-1}$ for many cases especially when the size of
$S$ is large. An alternative way is to approximate the solution of
(\ref{eqn:subprob}) with classical conjugate gradient (CG) decent
method. It is much faster than computing the exact solution. Besides CG, a better way is the preconditioned conjugate gradient (PCG) method \cite{saad2003iterative}.
The design of preconditioner is problem-dependent, which should be as close as possible to the system matrix $S$ and can be inversed efficiently.
Therefore, it is not an easy task to design a good preconditioner in general due to this tradeoff.
For signal/image reconstruction, such preconditioner has not been found in existing IRLS algorithms \cite{Chartrand08,Daubechies08,FOCUSS}.

By observing that $S$ is usually diagonally dominant in reconstruction problems, e.g. CS imaging, image inpainting and CS-MRI, we define a new preconditioner for best approximation in Frobenius norm $||\cdot||_F$:
\begin{eqnarray}
P = \arg \min_{X\in \mathcal{D}} ||S - X||_F,
 \label{eqn:ps}
\end{eqnarray}
where $\mathcal{D}$ denotes the class of diagonal or ``pseudo-diagonal" matrices. Here, the pseudo-diagonal matrix means a non-diagonal matrix whose inverse can be obtained efficiently like a diagonal matrix with $\mathcal{O}(N)$ time. Please note that the $G^TW^kG$ is always diagonal for any kind of $G$ in Fig. \ref{fig:G}. Due to the strong constraint, the possible diagonal or ``pseudo-diagonal" candidates for (\ref{eqn:ps}) are enumerable. In addition, we observe that $A^TA$ is often diagonally dominant in the image reconstruction problems. For example, in CS-MRI, $A = RF$ where $F$ denotes the Fourier transform and $R \in \mathbb{R}^{M\times N}$ ($M<N$) is a selection matrix containing
$M$ rows of the identity matrix. Therefore, $A^TA = F^TR^TRF$ is diagonally dominant as $R^TR$ is diagonal. For the image inpainting problem, $A^TA = R^TR$ is diagonal. This structure also holds when $A$ is a random projection matrix.

Based on this diagonally dominant effect,
it is not hard to find an accurate solution
\begin{eqnarray}
P =(\overline{A^TA} I + \lambda \Phi^T G^TW^k G\Phi),
 \label{eqn:pog}
\end{eqnarray}
where $\overline{A^TA}$ is the mean of diagonal elements of $A^TA$ and $I$ denotes the identity matrix. The preconditioning error in Frobenius norm $||S - P||_F$ is very small, due to diagonally dominant structure of $A^TA$.

As $A$ is known for the application, $\overline{A^TA}$ can be pre-estimated before the first iteration and is fixed for each iteration. Therefore in each iteration, $P^{-1} =\Phi^T (\overline{A^TA} I + \lambda G^TW^k G )^{-1}\Phi$ can be obtained with linear time.

Several advantages of the proposed preconditioner can be found when compared with existing ones \cite{Papandreou11,Lefkimmiatis12}.
To get the inverse, fast Fourier transforms are used in recent circulant preconditioners for image deblurring \cite{Papandreou11} \cite{Lefkimmiatis12}, while our model only requires linear time to obtain $P^{-1}$. Compared with conventional Jacobi preconditioner, we do not discard all non-diagonal information and therefore the preconditioner is more accurate. Moreover, our model can also handle the case when A or $\Phi$ is an operator, while other preconditioners \cite{Papandreou11,Lefkimmiatis12,rodriguez2009efficient} cannot because they require the exact values of $S$. 
Interestingly, the conventional Jacobi preconditioner can be derived by (\ref{eqn:ps}), when the original data is sparse (i.e. $\Phi = 1$) and $A$ is a numerical matrix.

 \begin{algorithm}[H]
\caption{FIRLS\_OG} \label{alg:FIRLS}
\begin{algorithmic}
   \STATE {\bfseries Input:} $A$,$b$,$x^1$, $G$, $\lambda$, $k=1$
   \WHILE{not meet the stopping criterion}
   \STATE Update $W^k$ by (\ref{eqn:W}) (\ref{eqn:gw})
   \STATE Update $S = A^TA+\lambda\Phi^TG^TW^k G \Phi$
   \STATE Update $P= \Phi^T ( \overline{A^TA}I + \lambda G^TW^k G )\Phi$, \\ \quad \quad \quad \ \ $P^{-1} =\Phi^T (\overline{A^TA} I + \lambda G^TW^k G )^{-1}\Phi$
   \WHILE{not meet the PCG stopping criterion}{
         \STATE Update $x^{k+1}$ by PCG for $Sx = A^Tb$ with preconditioner $P$
         }
   \ENDWHILE
   \STATE Update $k=k+1$
   \ENDWHILE
\end{algorithmic}
\end{algorithm}

Our method can be summarized in Algorithm
\ref{alg:FIRLS}. We denote this overlapping group sparsity version as FIRLS\_OG. Although our algorithm has double loops, we observe
that only $10$ to $30$ PCG iterations are sufficient to obtain a solution very
close to the optimal one for the problem (\ref{eqn:subprob}). In
each inner PCG iteration, the dominated cost is by applying $S$ and $P^{-1}$, which is denoted by $\mathcal{O}(\mathcal{C}_S+\mathcal{C}_P)$. When $A$ and $\Phi$ are dense matrices, $\mathcal{O}(\mathcal{C}_S+\mathcal{C}_P)= \mathcal{O}(N^2)$. When $A$ and $\Phi$ are the partial Fourier transform and wavelet transform in CS-MRI \cite{lustig2007sparse}, it is $\mathcal{O}(N \log N)$.

\subsection{Convergence analysis} \label{sec:conv}

\quad \textbf{Theorem 1}. \emph{The global optimal solution $x^*$ of (\ref{eqn:GWx}) is
the global optimal solution of original problem (\ref{eqn:probogs})}.

\vspace{0.2cm}
\textbf{Proof}. Suppose $x_1^*$ is the global optimal solution of (\ref{eqn:GWx}) and
$x_2^*$ is the global optimal solution of (\ref{eqn:probogs}). Consider $Q$ as a function corresponds to $x$ and $W$.
We have:
\begin{align}
&Q(x_1^*, W_1^*) \leq Q(x_2^*, W), \ \  \forall \ W; \\
&F(x_2^*) \leq F(x_1^*).
 \label{eqn:dw3}
\end{align}
Based on the inequalities (\ref{eqn:gx})(\ref{eqn:g21}), we have
\begin{align}
F(x) &\leq   Q(x, W^k)  \ \  \forall \ x;  \label{eqn:fxq}\\
F(x^k) &=   Q(x^k, W^k).
 \label{eqn:fxkq}
\end{align}
Therefore,
\begin{align}
F(x_2^*) \leq  F(x_1^*) = Q(x_1^*, W_1^*) \leq Q(x_2^*, W_2^*) =
F(x_2^*),
 \label{eqn:dw3}
\end{align}
which indicates $F(x_1^*) =  F(x_2^*)$. Here $W_1^*$ and $W_1^*$ are
weights of $x_1^*$, $x_2^*$ based on (\ref{eqn:gw}) and (\ref{eqn:Wog}).

\vspace{0.2cm}
\textbf{Theorem 2}. \emph{$F(x^k)$ is monotonically decreased by Algorithm 2, i.e.
$F(x^{k+1})\leq F(x^{k})$. In particular, we have $\lim_{k\rightarrow
\infty} (F(x^k) - F(x^{k+1})) = 0$.}

\vspace{0.2cm}
\textbf{Proof}. 
With the property (\ref{eqn:fxq}), we have
\begin{eqnarray}
F(x^{k+1}) \leq Q(x^{k+1},W^{k}).
 \label{eqn:conv1}
\end{eqnarray}
To balance the cost and accuracy when solving (15), we
apply the PCG method to decrease $Q(x,W^{k})$ and efficiently
obtain the solution $x^{k+1}$. Because $x^{k}$ is the initial guess for $x^{k+1}$, based on the property of PCG we have:
\begin{eqnarray}
Q(x^{k+1},W^{k}) \leq Q(x^{k},W^{k}).
 \label{eqn:conv2}
\end{eqnarray} \vspace{-0.0cm}
And we finally get:
\begin{eqnarray}
 F(x^{k+1})\leq Q(x^{k+1},W^{k}) \leq Q(x^{k},W^{k}) = F(x^{k}).
 \label{eqn:conv3}
\end{eqnarray} \vspace{-0.0cm}

$F(x)$ is convex and bounded. Due to the monotone convergence theorem, we have:
\begin{eqnarray}
\lim_{k\rightarrow \infty} (F(x^k) - F(x^{k+1})) = 0.
\label{eqn:conv3}
\end{eqnarray} \vspace{-0.0cm}

\vspace{0.2cm}
\textbf{Theorem 3}. \emph{
Any accumulation point of $\{x^k\}$ is a
stationary point of problem (\ref{eqn:GWx})}.

\vspace{0.2cm}
\textbf{Proof}. 
When we have any accumulation point $x^k = x^{k+1}$ for $k \rightarrow \infty$, it demonstrates the inner PCG
loop has converged for problem (\ref{eqn:subprob}). Therefore, it indicates
\begin{eqnarray}
(A^TA+\lambda\Phi^TG^TW^kG\Phi)x^{k} - A^Tb = 0.
 \label{eqn:xsubprob}
\end{eqnarray}
Consider $Q$ as a function corresponding to $x$ and $W$. We have
\begin{eqnarray}
\frac{\partial Q(x^k, W^k)}{\partial x} = 0.
 \label{eqn:xor}
\end{eqnarray}
In addition,
\begin{align}
&\frac{\partial Q(x^k, W^k)}{\partial W} \notag \\ = &\partial \{
\frac{\lambda}{2}(x^k)^T \Phi^T G^TWG\Phi x^k + \frac{\lambda}{2}
\sum_{i=1}^s \frac{1}{W^k_{g_i}} \}/\partial W \notag \\
=  &\frac{\lambda}{2}[(x^k)^T \Phi^T G^TG\Phi x^k -
\sum_{i=1}^s(W^k_{g_i})^{-2}].
 \label{eqn:dw}
\end{align}
Based on (\ref{eqn:gw}) and (\ref{eqn:Wog}), it can be
rewritten as:
\begin{align}
&\frac{\lambda}{2}\sum_{i=1}^s  [||(G\Phi x^k)_{g_i}||_2^2 - (
||(G\Phi x^k)_{g_i}||_2^2 + \epsilon)] 
& = \lambda s\epsilon/2.
 \label{eqn:dw2}
\end{align}
Note that $\epsilon$ is negligible and we finally have
\begin{align}
\frac{\partial Q(x^k, W^k)}{\partial x} = \frac{\partial Q(x^k,
W^k)}{\partial W} \approx 0.
 \label{eqn:dw3}
\end{align}
Hence $x^k$ is a stationary point of (\ref{eqn:GWx}) when ${k\rightarrow \infty}$.

It indicates that the algorithm converges to a local minimum of the problem. In our experiments, if we let the initial guess $x^0 = A^Tb$, an accurate solution can be always obtained. Note that Theorems 1, 2 and 3 always hold no matter how many inner PCG iterations are used.

\section{FIRLS for Total Variation}

We have presented an efficient algorithm for overlapping group sparsity under an orthogonal sparse basis $\Phi$. In image reconstruction problems, another widely used sparsity regularizer is the TV. Due to the non-orthogonality of the TV semi-norm, the FIRLS\_OG algorithm can not be applied to solve the TV problem.
In this section, we will present an efficient algorithm for TV
based image reconstruction. For brevity and clarity, we first
present the algorithm for single channel image reconstruction and then extended
it to multi-channel reconstruction \cite{bresson2008fast}. 

\subsection{An alternative formulation for total variation}
TV minimization exploits the sparsity of the image in the gradient
domain. 
For brevity, we assume the image is $n$ by $n$ with $n \times n = N$.
Let $D_1$, $D_2$ be two $N$-by-$N$ two first-order finite difference matrices in vertical and
horizontal directions.
\begin{eqnarray}
D_1 = \begin{bmatrix}
1 &  &  & &\\
-1 & 1  &  & &\\
 & -1 & 1 & &\\
 &  &...  & ... &\\
 &    & & ...  & ...\\
 &  & &  & -1 & 1 & \\
& & & &  & -1 & 1
\end{bmatrix},
 \label{eqn:uncL1}
\end{eqnarray} \vspace{-0.0cm}
\begin{eqnarray}
\quad D_2 = \begin{bmatrix}
1 &  &  & &\\
... & ...  &  & &\\
-1 & ... & 1 & &\\
 & -1 &...  & 1 &\\
 &    &...  &  & ...\\
 &  & & -1 & ... & 1 & \\
& & & & -1 & ... & 1
\end{bmatrix}.
 \label{eqn:uncL1}
\end{eqnarray} \vspace{-0.0cm}
The main diagonal elements of $D_1$ and $D_2$ are all ones. The first diagonal elements below the main diagonal are all minus ones in $D_1$, while
 in $D_2$ $n$-th diagonal elements below the main diagonal are all minus ones.
 With these notations, the $\ell_1$ and isotropic
 TV based image reconstruction can be reformulated as:
\begin{eqnarray}
\min_x \{\frac{1}{2}||Ax-b||_2^2 + \lambda ||D_1 x||_{1} + \lambda ||D_2 x||_{1} \},
 \label{eqn:l1}
\end{eqnarray}
and
\begin{eqnarray}
\min_x \{F(x) = \frac{1}{2}||Ax-b||_2^2 + \lambda ||[D_1 x,D_2 x]||_{2,1} \}.
 \label{eqn:iso}
\end{eqnarray}
Here, the $\ell_{2,1}$ norm is the summation of the $\ell_{2}$ norm
of each row, which is a special case of (\ref{eqn:l21}). Here and later, we denote $[\ ,\ ]$  as the concatenating of
the matrices horizontally. To avoid repetition, all the following
derivations only consider isotropic TV function (\ref{eqn:iso}). $\ell_1$-based TV function can be derived in the same
way.

Considering the Young's inequality in (\ref{eqn:gx}), 
 we majorize (\ref{eqn:iso}) by the MM method \cite{Hunter04}:
\begin{eqnarray}
Q(x,W^{k}) = \frac{1}{2}||Ax-b||_2^2+\frac{\lambda}{2}x^T  D_{1}^TW^kD_{1}x \notag \\
 + \frac{\lambda}{2}x^T  D_{2}^TW^kD_{2}x + \frac{\lambda}{2}\mathrm{Tr}((W^k)^{-1}),
 \label{eqn:Gx}
\end{eqnarray}
where $\mathrm{Tr}()$ denotes the trace. $W^k$ is a diagonal weight matrix in the
$k$-th iteration:
\begin{eqnarray}
W_{i}^k = 1/\sqrt{(\nabla_{1}{x}^k_{i})^2+(\nabla_{2}{x}^k_{i})^2 + \epsilon}, \ i = 1,2,..., N,
\label{eqn:wi}
\end{eqnarray}
and
\begin{eqnarray}
W^{k} =\begin{Bmatrix}
W^k_{1}  &  &  &   \\
 &  W^k_{2} &  &  \\
  &  & ...  &  \\
 &  &  & W^k_{N} \\
\end{Bmatrix}.
\label{eqn:W}
\end{eqnarray}
When $D_1 = D_2 = I$, it is identical to the $\ell_1$ norm minimization as in the conventional IRLS methods \cite{Chartrand08,Daubechies08,FOCUSS}.



\subsection{Accelerating with PCG and incomplete LU decomposition}
After the weight matrix is updated by (\ref{eqn:wi}) and (\ref{eqn:W}), the problem is to update $x$. With the same rule as that in the overlapping group sparsity regularization, we have
\begin{eqnarray}
(A^TA+\lambda D_{1}^TW^kD_{1} + \lambda D_{2}^TW^kD_{2})x = A^Tb.
 \label{eqn:sub}
\end{eqnarray}
Similar to (\ref{eqn:subprob}), the system matrix here is in large scale. We have discussed that the system matrix is not dense but follows some special structure in image reconstruction.
A good solver should consider such special structure of the problem.
In TV based image deblurring problems, by observing that $A$ has a circulant structure (under periodic boundary conditions), many efficient algorithms have been proposed to accelerate the minimization \cite{Lefkimmiatis12,yang2009fast,chan2011augmented}. However, these algorithms can not be applied to the TV reconstruction problems.

Based on the diagonally dominant prior information in image reconstruction, we can obtained an accurate preconditioner like (\ref{eqn:pog}).
\begin{eqnarray}
P = \overline{A^TA} I + \lambda D_{1}^TW^kD_{1} + \lambda D_{2}^TW^kD_{2}
 \label{eqn:ptv}
\end{eqnarray}
%
However, the inverse can not be efficiently obtained for this preconditioner, due to the non-orthogonality of $D_1$ and $D_2$.
Fortunately, $P$ has a sparse structure due to the forms of $D_1$ and $D_2$. We observe that preconditioner $P$ in (\ref{eqn:ptv}) is a penta-diagonal matrix \cite{saad2003iterative} and is symmetric, which has the following form:
\begin{eqnarray}
P = \begin{bmatrix}
a_1 & b_1   &  & c_1 &  &  & \\
b_1 & a_2   & b_2  &  & c_2 &  & \\
 & b_2 & a_3  & ... &  & ... & \\
 & & ...  & ... &  &  &  c_{N-n}\\
c_1 &    &  &  &  &  & \\
 & c_2   &  &  &  & ... & \\
 &  & ...   & & ... & ...  & b_{N-1}\\
 &  &  &c_{N-n}  &   &b_{N-1} & a_{N}
\end{bmatrix}.
\end{eqnarray}
Such penta-diagonal matrix has incomplete LU decomposition $P\thickapprox LU$, where
\begin{eqnarray}
L = \begin{bmatrix}
1 &  &  &  &  &  &  & \\
\frac{b_1}{a_1} & 1 &   &  &   &  & \\
 & \frac{b_2}{a_2} & 1 &  &    &  & \\
 & & ... & ... &  &  &  &  \\
\frac{c_{1}}{a_{1}}  &  &  & \frac{b_n}{a_n} & 1 &  & \\
   & ... &  & & ... & ...  & \\
   &  & \frac{c_{N-n}}{a_{N-n}} & ... &  & \frac{b_{N-1}}{a_{N-1}} &1
\end{bmatrix},
\\
U= \begin{bmatrix}
a_1 & b_1 & ... & c_1 &  &  &  & \\
 & a_2 & b_2  & ... &  c_2 & &  & \\
 &   & ... & ... &  & ... &  \\
  &   &  &a_{N-n}  & b_{N-n} & ... & c_{N-n} \\
 &   &  & & ... & ...  & \\
  &   &  & &  & a_{N}  & b_{N-1}\\
 &   &  &  &  & & a_{N}
\end{bmatrix}.
\end{eqnarray}
The decomposition is very accurate as $P$ is diagonally dominated
with $a_i\gg b_i^2, a_i\gg c_i^2$ and $a_i\gg b_i c_i$ for all $i$. To the best of our knowledge, this incomplete LU decomposition is first proposed for TV minimization.
Due to the special structure of $P$, the incomplete LU decomposition only takes $\mathcal{O}(N)$ time. Therefore, the total time to obtain $P^{-1} \thickapprox U^{-1}L^{-1}$ is $\mathcal{O}(N)$. We can conclude the proposed method for TV reconstruction in Algorithm
\ref{alg:FIRLSTV}. The convergence of this algorithm can be proven in a similar way as Section \ref{sec:conv}, which is omitted to avoid repetition.

 \begin{algorithm}[H]
\caption{FIRLS\_TV} \label{alg:FIRLSTV}
\begin{algorithmic}
   \STATE {\bfseries Input:} $A$, $b$, $x^1$, $\lambda$, $k=1$
   \WHILE{not meet the stopping criterion}
   \STATE Update $W^k$ by (\ref{eqn:wi}) and (\ref{eqn:W})
   \STATE Update $S = A^TA+\lambda D_{1}^TW^kD_{1} + \lambda D_{2}^TW^kD_{2}$
   \STATE Update $P = \overline{A^TA}I +\lambda D_{1}^TW^kD_{1} + \lambda D_{2}^TW^kD_{2} \thickapprox LU$, $P^{-1} \thickapprox U^{-1}L^{-1}$
   \WHILE{not meet the PCG stopping criterion}{
         \STATE Update $x^{k+1}$ by PCG for $Sx = A^Tb$ with preconditioner $P$
         }
   \ENDWHILE
   \STATE Update $k=k+1$
   \ENDWHILE
\end{algorithmic}
\end{algorithm}

\section{Experiments}

\subsection{Experiment setup}


The experiments are conducted using MATLAB on a desktop computer with 3.4GHz Intel core i7 3770 CPU.
We validate different versions of our method on wavelet tree sparsity based reconstruction, wavelet joint sparsity reconstruction, TV and JTV reconstruction.  To avoid confusion, we denote the tree sparsity version as FIRLS\_OG and non-overlapping joint sparsity version FIRLS\_MT.
The version for standard $\ell_1$ norm minimization is denoted by FIRLS\_L1.
FIRLS\_TV and FIRLS\_JTV denotes the TV and JTV reconstruction, respectively.

Note that some algorithms need a very small number of iterations to converge (higher convergence rate), while they cost more time in each iteration (higher complexity). The others take less time in each iteration; however, more iterations are required.  As we are interested in fast reconstruction, an algorithm is said to be better if it can achieve higher reconstruction accuracy with less computational time.
\subsection{The accuracy of the proposed preconditioner}

\vspace{-0.0cm}
\begin{figure}[htbp]
\centering \vspace{-0.0cm}
 \includegraphics[scale=0.32]{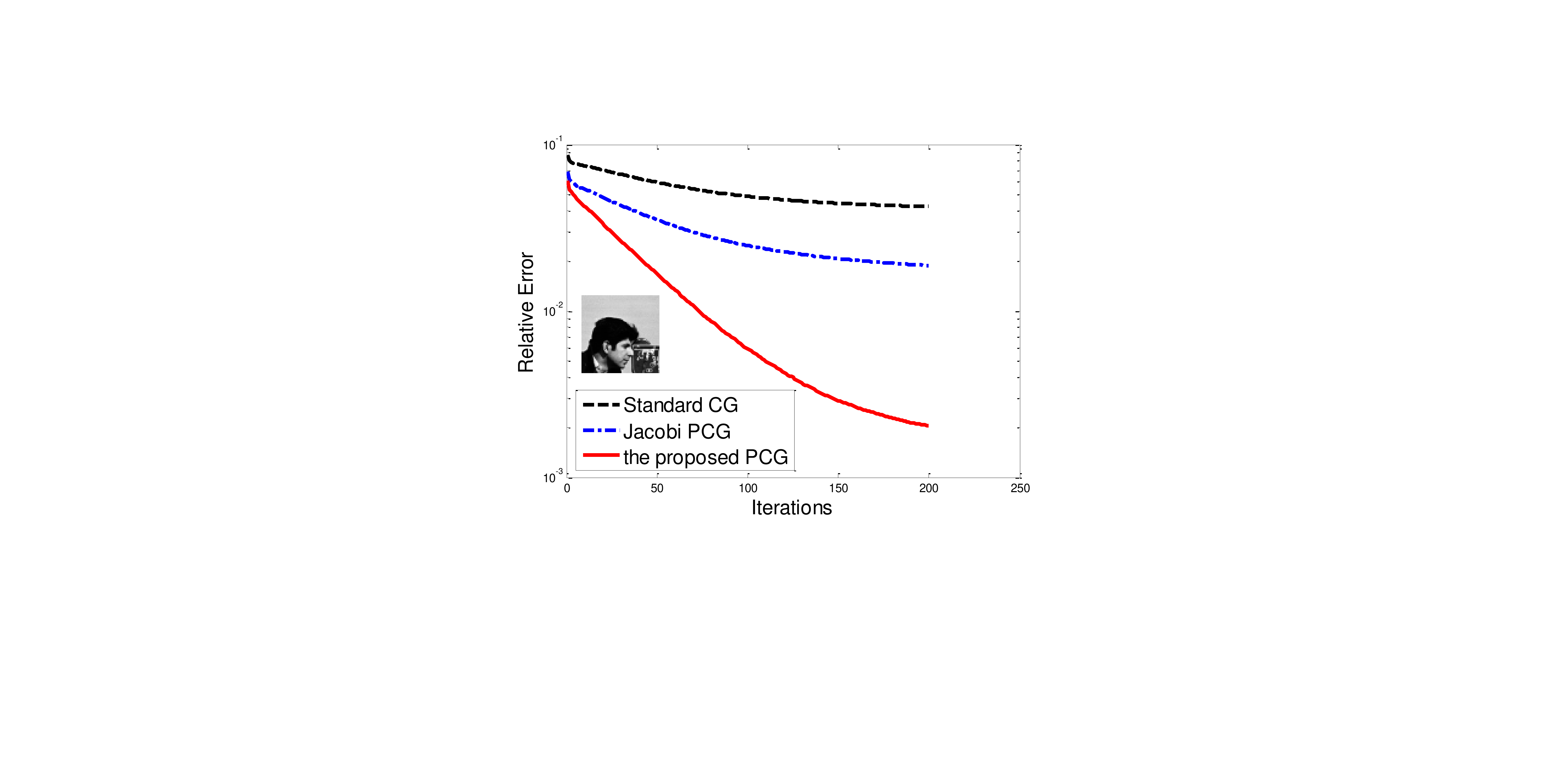}
\caption{Convergence rate comparison among standard CG, Jacobi PCG and the proposed PCG for $\ell_1$ norm minimization.}\label{fig:PCG}\vspace{-0.0cm}
\end{figure} \vspace{-0.0cm}

\begin{table*}[t]
\caption{Computational cost comparison between FOCUSS \cite{FOCUSS} and the proposed method} \label{vsFOCUSS} 
\centering
\begin{tabular}{lcccccccr}
\hline
          &\vline&  &            FOCUSS \cite{FOCUSS} & &\vline&     &  FIRLS\_L1  & \\
\hline
Time (seconds)   &\vline &64.8  &110.8  &727.7              &\vline& 10.5   &29.8   &{120.2} \\
MSE  &\vline&0.0485   &0.0442   &0.0432            &\vline& 0.0481 &0.0440  &0.0427 \\
\hline
\end{tabular}
\end{table*}

One of our contributions is the proposed pseudo-diagonal preconditioner for sparse recovery. First, we conduct an experiment to validate its effectiveness with the orthogonal wavelet basis.
Without loss of generality, a patch ($64\times 64$) cropped from the cameraman image is used for reconstruction, which is feasible to obtain the closed form solution of $S^{-1}$ for evaluation. As most existing preconditioners cannot support the inverse of operators, the sampling matrix is set as the random projection and $\Phi$ is a dense matrix for wavelet basis here. Fig. \ref{fig:PCG} demonstrates the performance of the proposed PCG compared with Jacobi PCG and standard CG for the problem (\ref{eqn:subprob}). The performance of the proposed PCG with less than 50 iterations is better than that of CG and Jacobi PCG with 200 iterations. Although Jacobi preconditioner is diagonal, it removes all the non-diagonal elements which makes the preconditioner less precise.

\begin{figure}[htbp]
\centering \vspace{-0.0cm}
        \includegraphics[scale=0.32]{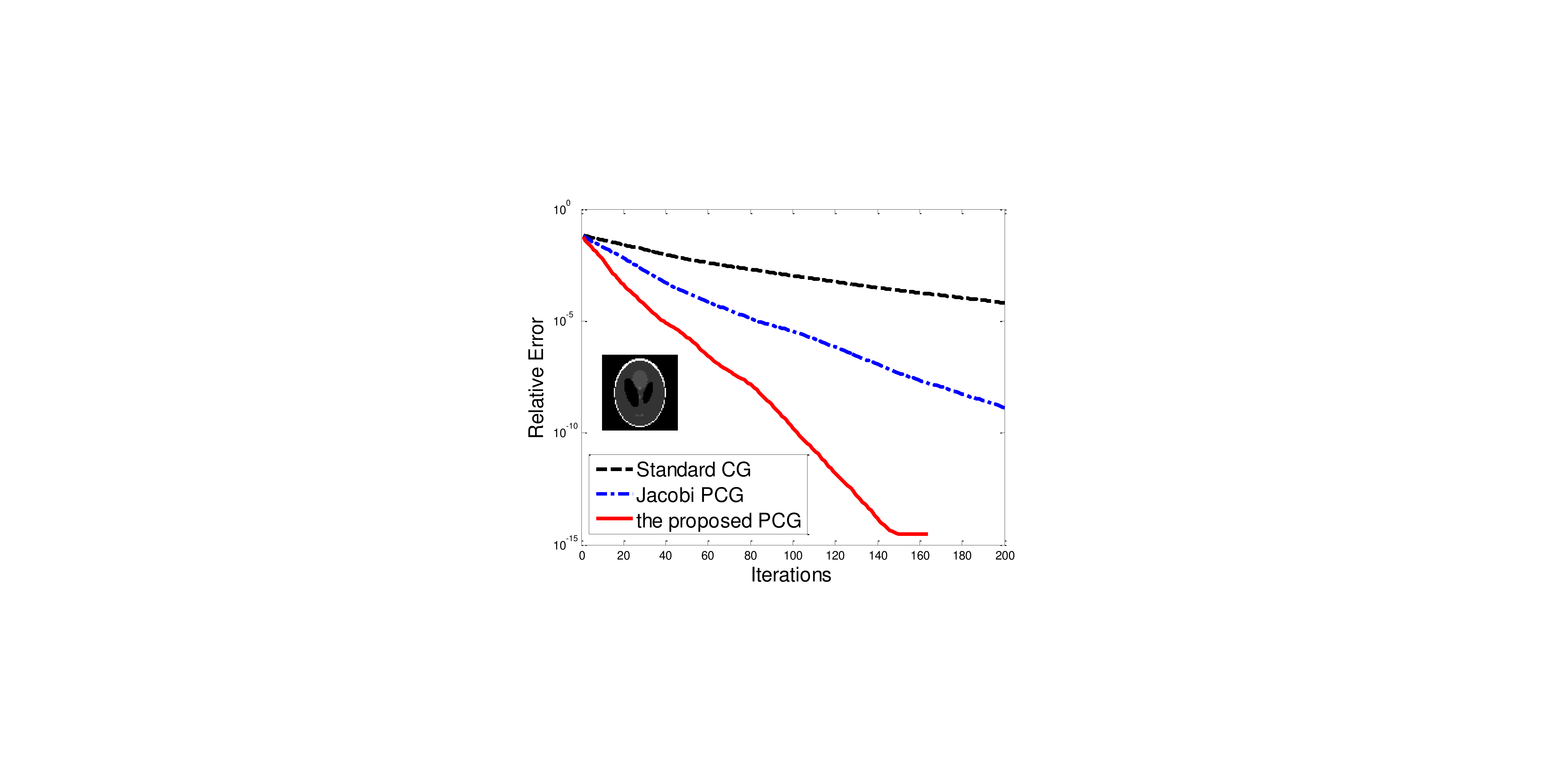}
\caption{Convergence rate comparison among standard CG, Jacobi PCG and the proposed PCG for TV minimization.}
\label{fig:tvrec}\vspace{-0.0cm}
\end{figure}

To validate the effectiveness of the proposed
preconditioner in TV reconstruction, we take experiments on the Shepp-Logan phantom image
with $64\times 64$ pixels. The Shepp-Logan phantom image is very smooth and is an ideal example to validate TV reconstruction.
The relative errors of CG, PCG Jacobi and the proposed method are shown in Fig. \ref{fig:tvrec}.
It shows that only 20 iterations of PCG with the proposed
preconditioner can outperform conventioanal CG with 200 iterations.
Jacobi PCG requires approximately 2 times of iterations to reach the
same accuracy as our method, because it discards all non-diagonal
information directly and makes the preconditioning less precise. Comparing with the results in Fig. \ref{fig:PCG}, our preconditioner seems less powerful on TV reconstruction.  This is expected as we further decompose the preconditioner into two triangle matrices L and U, which introduces minor approximation error. However, it still converges much faster than the existing Jacobi PCG. These experiments demonstrate that the inner loop subproblem in our method is solved efficiently with the proposed preconditioner.

\subsection{Convergence Rate and Computational Complexity}

One of the properties of the proposed FIRLS is its fast convergence rate, i.e., only a small number of iterations can achieve high reconstruction accuracy. In addition, each iteration has low computational cost. To validate its fast convergence rate, we compare it with three existing algorithms with known convergence rate. They are the IST algorithm SpaRSA \cite{Wright09}, FISTA \cite{Beck09} and IRLS algorithm FOCUSS \cite{FOCUSS}, with $\mathcal{O}(1/k)$,  $\mathcal{O}(1/k^2)$ and exponential convergence rates, respectively. Mean squared error (MSE) is used as the evaluation metric.

\begin{figure}[htbp]
\centering
 \includegraphics[scale=0.42]{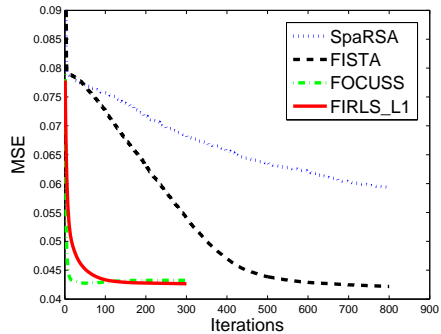}
\caption{Convergence Rate Comparison among FOCUSS, FISTA and SpaRSA for $\ell_1$ norm minimization. }\label{fig:conv}\vspace{-0.0cm}
\end{figure}

 The test data is a random 1D signal of length 4000, with $10\%$ elements being non-zeros. The number of measurements are $800$. Fig. \ref{fig:conv} demonstrates the comparison. In each iteration, FOCUSS needs to compute the inverse of a large-scale matrix, and the proposed method uses 30 PCG iterations to approximate the inverse. Both FOCUSS and the proposed method converge within 200 iterations. FISTA tends to converge at about 800 iterations. However, SpaRSA requires much more than 800 iterations to converge. 
 Table \ref{vsFOCUSS} lists the reconstruction results at different CPU time between FOCUSS and the proposed method. The proposed algorithm always achieves more accurate result in much less time.  After convergence, the 0.0005 difference in terms of MSE may be caused by approximation or rounding errors. With the size of the data becomes larger, the time cost of FOCUSS will increase at a cubic speed. More importantly, it is not known how to solve the overlapping group sparsity problem with FOCUSS.

\subsection{Application: compressive sensing MRI}

Compressive sensing MRI (CS-MRI) \cite{lustig2007sparse} is one of the most successful applications of compressive sensing and sparsity regularization.
There are various sparsity patterns on MR images. Therefore, we validate the performance of different versions of our method on CS-MRI as a concrete reconstruction instance.
Partial but not full k-space data is acquired and the final MR image can be reconstructed by exploiting the sparsity of the image. With little information loss, this scheme can significantly accelerate MRI acquisition.
In CS-MRI, $A=RF$ is an undersampled Fourier operator, where $F$ is the Fourier transform and $R \in \mathbb{R}^{M\times N}$ is a selection matrix containing
$M$ rows of the identity matrix. Therefore, $A^TA=F^TR^TRF$ is diagonally dominant as $R^TR$ is diagonal. Based on (\ref{eqn:ps}), $\overline{A^TA}$ is identical to the sampling ratio (a fixed scalar).



Following previous works, Signal-to-Noise Ratio (SNR) are used as metric for result
evaluation:
\begin{eqnarray}
SNR=10\log_{10}(V_{s}/V_{n}),
\end{eqnarray}
where $V_{n}$ is the Mean Square Error between the original image
$x_{0}$ and the reconstructed $x$; $V_{s}=var(x_{0})$ denotes the
variance of the values in $x_{0}$.





\subsection{CS-MRI}

\subsubsection{CS-MRI with wavelet tree sparsity}

\begin{figure}[htbp]
\centering
 \includegraphics[scale=0.3]{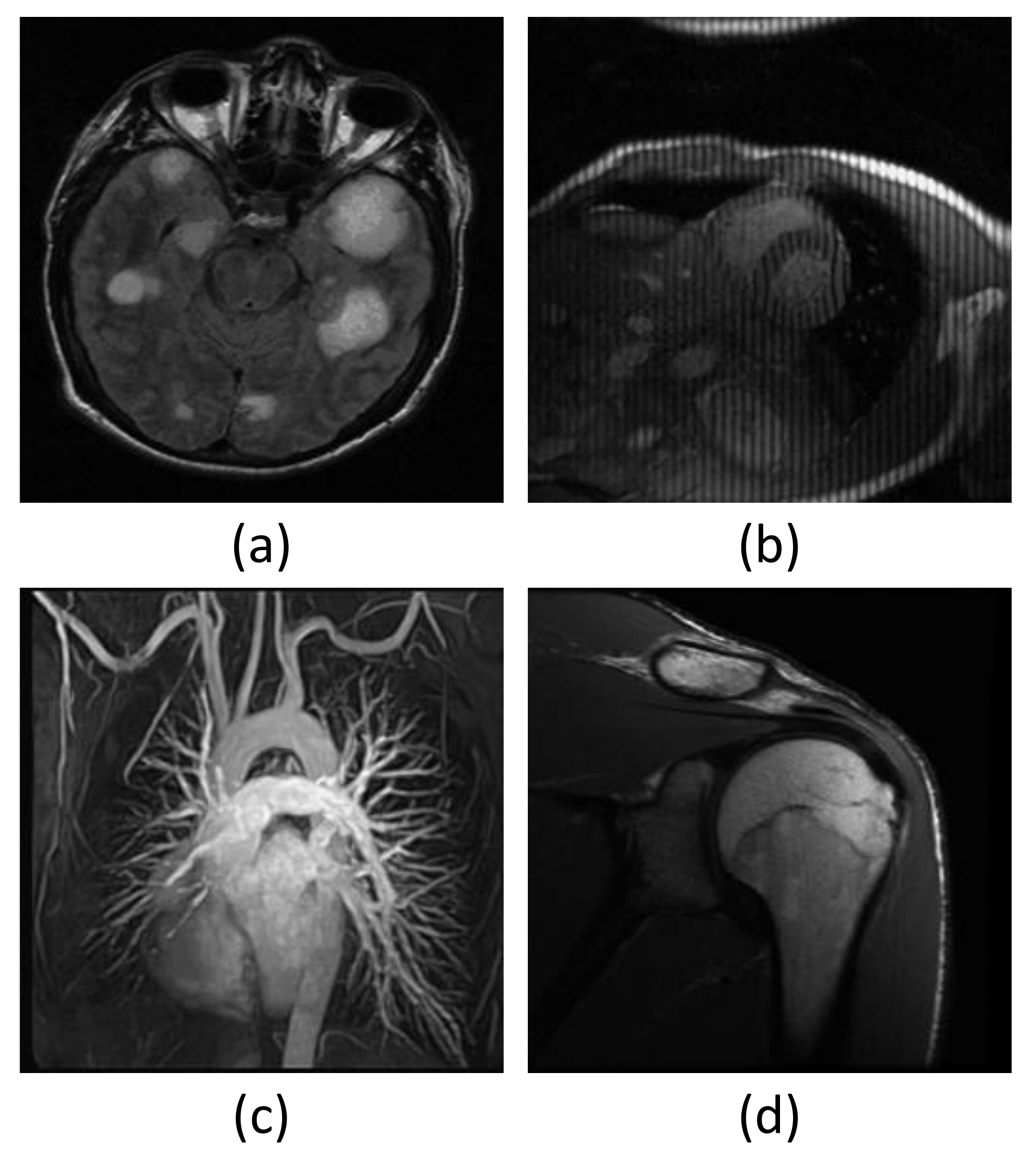}
\caption{ The original images: (a) Brain; (b) Cardiac; (c) Chest; (d) Shoulder.}\label{fig:ori}
\end{figure}
\begin{figure*}[t]
\centering
 \includegraphics[scale=0.4]{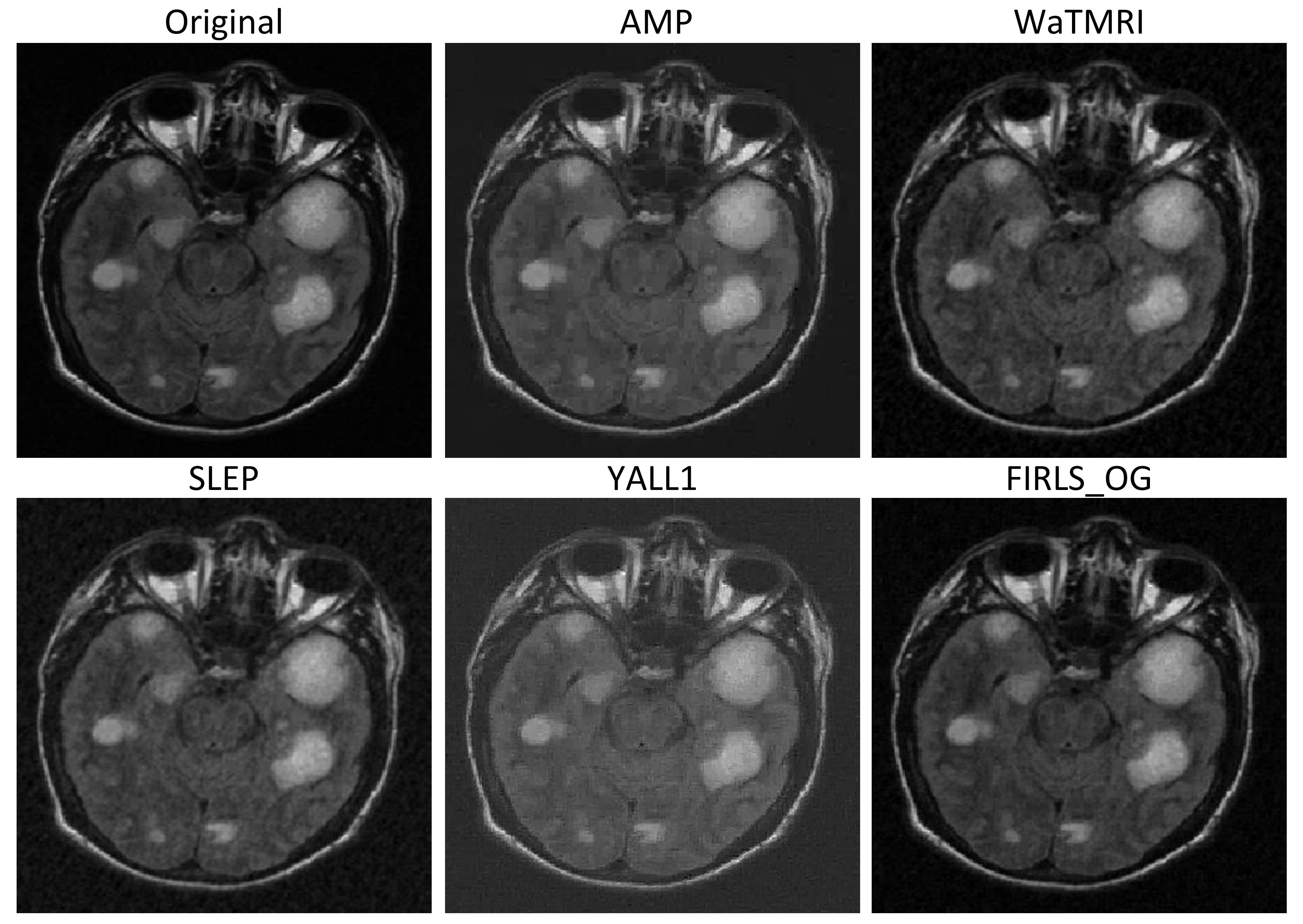}
\caption{Visual comparison on the Brain image with $25\%$ sampling. The SNRs of AMP \cite{AMP}, WaTMRI \cite{chen2012compressive}, SLEP \cite{SLEP}, YALL1 \cite{Deng11} and the proposed method are 15.91, 16.72, 16.49, 12.86 and 18.39, respectively.}\label{fig:treeVisual}\vspace{-0.0cm}
\end{figure*}
The MR images are often piecewise smooth, which are widely assumed to be sparse under the wavelet basis or in the gradient domain \cite{lustig2007sparse,ma2008efficient,yang2010fast,huang2011efficient}.
Furthermore, the wavelet coefficients of a natural image yield a quadtree. If a coefficient is zero or nonzero, its parent coefficient also tends to be zero or nonzero. This wavelet tree structure has already been successfully utilized in MR image reconstruction, approximated by the overlapping group sparsity \cite{chen2012compressive,chen2013benefit}.
Tree-structured CS-MRI method \cite{chen2012compressive,chen2013benefit} has been shown to be superior to standard CS-MRI methods \cite{lustig2007sparse,ma2008efficient,huang2011efficient}.
Therefore, we compare our algorithm with two latest and fastest tree-based algorithms, turbo AMP \cite{AMP} and WaTMRI \cite{chen2012compressive}. In addition, overlapping group sparsity solvers SLEP \cite{SLEP,yuan2013efficient} and YALL1 \cite{Deng11} are also compared. The total number of iterations is 100 except that turbo AMP only runs 10 iterations due to its higher time complexity. Followed by the previous works \cite{ma2008efficient,huang2011efficient,chen2012compressive}, four MR images with the same size $256\times 256$ are used for testing, which are shown in Fig. \ref{fig:ori}. Using a similar sampling strategy, we randomly choose more Fourier coefficients from low frequency and less on high frequency. The sampling ratio is defined as the number of sampled measurements divided by the total size of the signal/image.


A visual comparison on the Brain image is shown in Fig. \ref{fig:treeVisual}, with a $25\%$ sampling ratio. Visible artifacts can be found on the results by YALL1 \cite{Deng11}. The image reconstructed by the AMP \cite{AMP} tends to be blurry when compared with the original. The image recovered by SLEP \cite{SLEP} is noisy.
Our method and WaTMRI \cite{chen2012compressive} produce the most accurate results in terms of SNR. 
Note that WaTMRI has more parameters required to be tuned due to its variable splitting strategy.
Besides SNR, we also compare the mean structural similarity \cite{wang2004image} (MSSIM) of different images, which mimics the human visual system. The MSSIM for the images recovered by AMP \cite{AMP}, WaTMRI \cite{chen2012compressive}, SLEP \cite{SLEP}, YALL1 \cite{Deng11} and the proposed method are 0.8890, 0.8654, 0.8561, 0.7857 and 0.9009. In terms of MSSIM, our method still has the best performance, which is consistent with the observation in terms of SNR.



\begin{figure}[htbp]
\centering
 \includegraphics[scale=0.24]{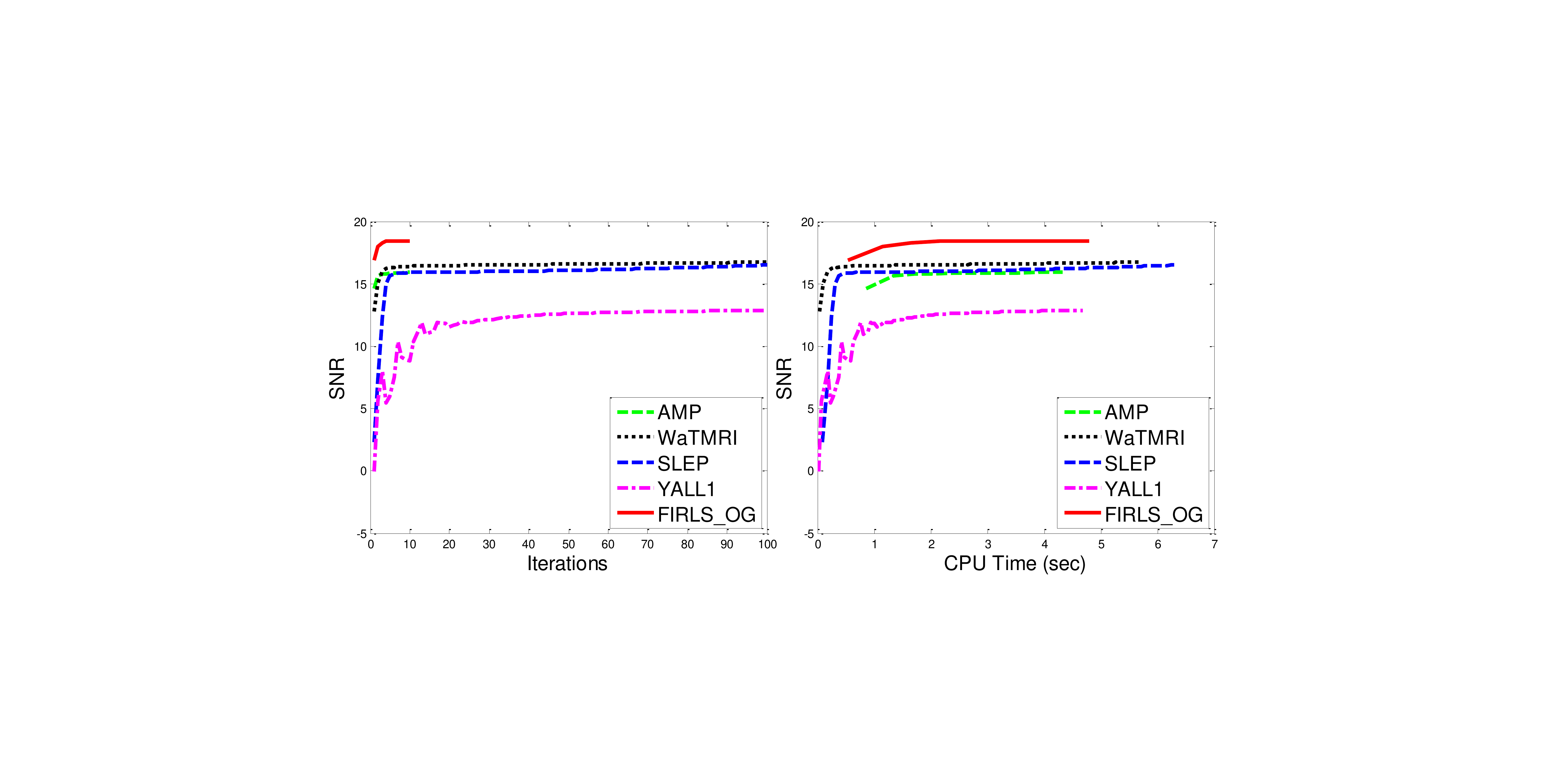}
\caption{Convergence speed comparison on the Brain image with $25\%$ sampling. Left: SNR vs outer loop iterations. Right: SNR vs CPU time. The SNRs of reconstructed images with these algorithms are 15.91, 16.72, 16.49, 12.86 and 18.39 respectively. The time costs are 4.34 seconds, 5.73 seconds, 6.28 seconds, 4.71 seconds and 4.80 seconds, respectively.}\label{fig:treeP}
\end{figure}

The corresponding convergence speed of the this experiment is presented in Fig. \ref{fig:treeP}. From SNR versus outer loop iterations, the proposed algorithm far exceeds that of all other algorithms, which is due to the fast convergence rate of IRLS. However, there is no known convergence rate  better than $\mathcal{O}(1/k^2)$ for WaTMRI and SLEP, and $\mathcal{O}(1/k)$ for YALL1, respectively. These results are consistent with that in previous work \cite{chen2012compressive}. For the same number of total iterations, the computational cost of our method is comparable to the fastest one YALL1, and it significantly outperforms YALL1 in terms of reconstruction accuracy. SLEP has the same formulation as ours. To reach our result in this experiment, it requires around 500 iterations with about 43 seconds. Similar results can be obtained on the other testing images. The results on the four images with different sampling ratios are listed in Table 2. Our results are consistently more accurate.

\begin{table}[htbp]
\caption{Average SNR (dB) comparisons on the four MR images with wavelet tree sparsity.}
\label{sampleall}
\centering
\begin{tabular}{cccccc}
\hline
\hline
Sampling Ratio & $20\%$ &$23\%$ &$25\%$ &$28\%$ &$30\%$ \\
\hline
AMP \cite{AMP} &11.64	&15.7	&16.43	&17.08	&17.44 \\
WaTMRI \cite{chen2012compressive} &15.56	&17.43	&18.23	&19.22	&20.45\\
SLEP \cite{SLEP} &11.59	&16.51	&17.36	&18.51	&20.07  \\
YALL1 \cite{Deng11} &12.13	&13.29	&14.12	&15.29	&16.07 \\
FIRLS\_OG  &\textbf{15.67}	&\textbf{18.78}	&\textbf{19.43}	&\textbf{20.53}	&\textbf{21.52}        \\
\hline
\hline
\end{tabular}
\end{table}

\subsubsection{CS-MRI by TV reconstruction}

TV is another popular regularizer for MRI reconstruction and the images recovered by TV tend to be less noisy \cite{lustig2007sparse}.
For TV based reconstruction, we compare our method with classical method CG \cite{lustig2007sparse} and the fastest ones TVCMRI \cite{ma2008efficient}, RecPF \cite{yang2010fast}, FCSA \cite{huang2011efficient} and SALSA \cite{afonso2010fast}.

\begin{figure}[htbp]
\centering
        \includegraphics[scale=0.4]{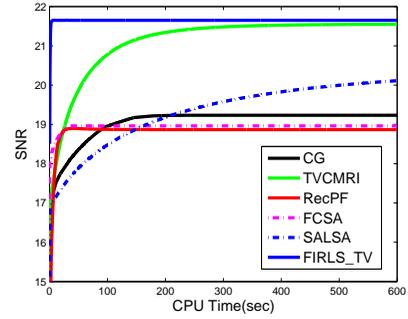}
\caption{Convergence rate comparison for TV minimization on the Chest image with $25\%$ sampling.}\label{fig:csrate}\vspace{-0.0cm}
\end{figure}

The convergence speed of different algorithms on the Chest image is presented in Fig. \ref{fig:csrate}. 
It is worthwhile to note that no closed form solutions exist for the subproblems of these algorithms. Therefore, the subproblems in these algorithms are often solved in an approximate way. Therefore, it is important to evaluate the accuracies of these algorithms.
From the figure, the final results of our method and TVCMRI are almost the same while the others converges to different results.
We further found that only TVCMRI has analyzed their global convergence (in Section 2.3 of \cite{ma2008efficient}), while the accuracy of all the other methods \cite{lustig2007sparse,yang2010fast,huang2011efficient,afonso2010fast} has not been discussed in details. For the four MR images, the average SNRs of CG \cite{lustig2007sparse}, TVCMRI \cite{ma2008efficient}, RecPF \cite{yang2010fast}, FCSA \cite{huang2011efficient}, SALSA \cite{afonso2010fast} and the proposed algorithm are 19.45, 21.78, 21.70, 21.53 21.95 and 23.07, respectively.



\begin{figure*}[htbp]
\centering \vspace{-0.0cm}
        \includegraphics[scale=0.39]{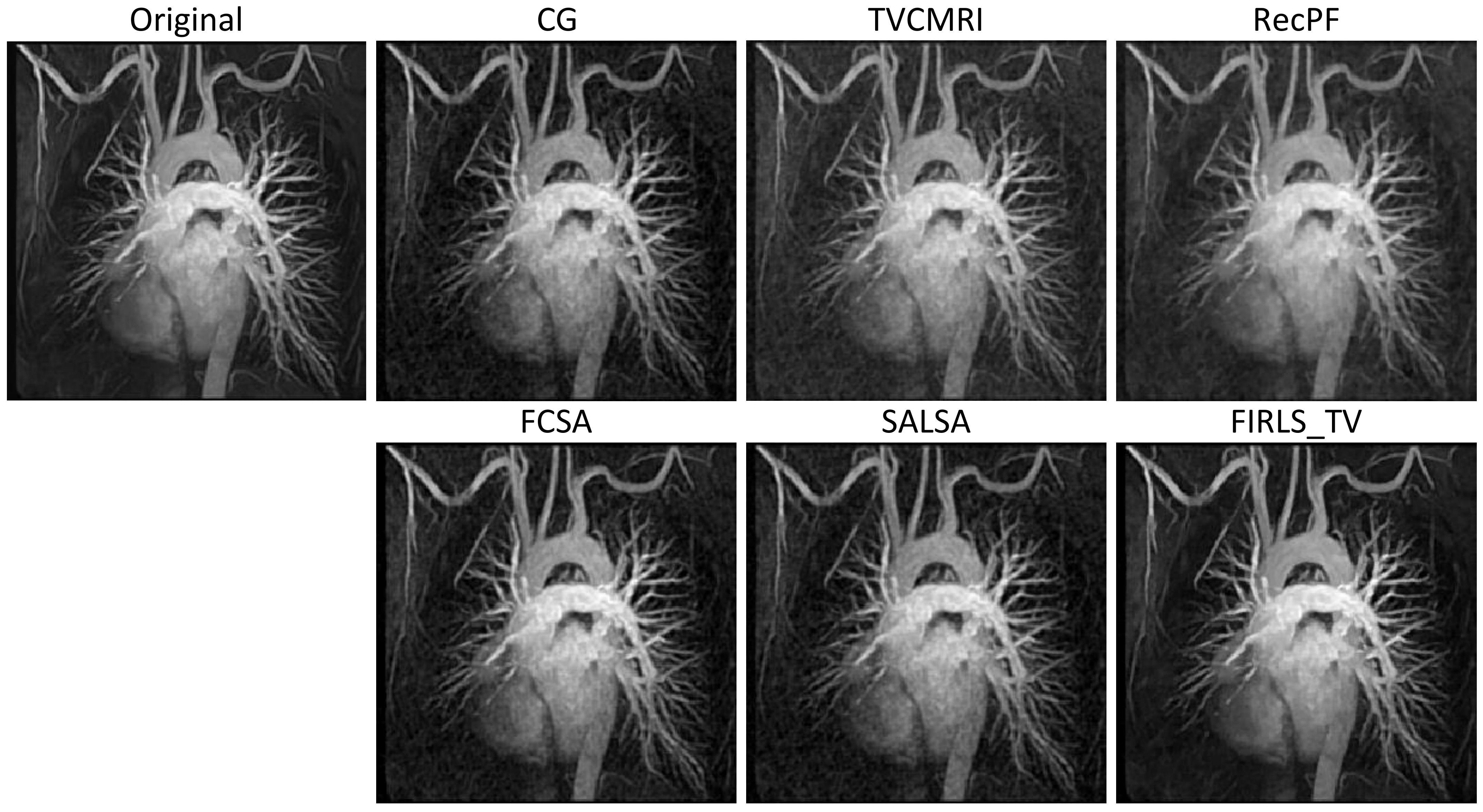}
\caption{Chest MR image reconstruction from $25\%$ sampling. All methods terminate after 4 seconds. The SNRs for CG, TVCMRI, RecPF, FCSA, SALSA and the proposed are 17.13, 17.32, 16.18, 18.28, 16.96 and 21.63, respectively.}\label{fig:visual}
\end{figure*}

We then terminate each algorithm after a fixed toleration is reached, e.g. $10^{-3}$ of the relative solution change.
The final SNR and convergence speed of different methods are listed in Table \ref{conspeedtv}. To produce a similar result of TVCMRI, our method only requires about its $1/70$ computational time. These convergence performances are not surprising. FIRLS is converges exponentially fast (as shown in Fig. \ref{fig:conv}) and require the fewest iterations.
FCSA is a FISTA based algorithm, which has $\mathcal{O}(1/k^2)$ convergence rate. It converges with the second fewest iterations. For the rest algorithms, there is no known convergence rate better than $\mathcal{O}(1/k)$.

\begin{table}[htbp]
\caption{Quantitative comparison of convergence speed on the Chest image by TV regularization with $25\%$ sampling.}
\label{conspeedtv}
\begin{tabular}{cccc}
\hline \hline
&Iterations   &CPU time (sec) &SNR (dB)
\\ \hline
CG \cite{lustig2007sparse} &3181 &397.8 &19.23 \\
TVCMRI \cite{ma2008efficient} &21392 &495.1 &21.54 \\
RecPF \cite{yang2010fast} &7974 &163.4 &18.86 \\
FCSA \cite{huang2011efficient} &1971 &39.6 &18.96\\
SALSA \cite{afonso2010fast} &9646 &882.4 &20.13 \\
FIRLS\_TV  &\textbf{29} &\textbf{6.9} &\textbf{21.65} \\
\hline  \hline
\end{tabular}
\end{table}

Due to the relatively slower convergence speed,
we note that previous methods \cite{lustig2007sparse,ma2008efficient,yang2010fast,huang2011efficient} often terminate after a fixed number of iterations (e.g. 200) in practice.
This is because the exactly convergence is time consuming that may not be feasible for clinic applications. Following by this scheme, we run TVCMRI 200 iterations. All the other algorithms terminate after the same running time of TVCMRI (i.e. around 4 seconds). The reconstruction results on the Chest MR image are shown in Fig. \ref{fig:visual}. A close look shows that our method preserve highest organ-to-background contrast without contaminated by reconstruction noise. Such results are expected if we take a review on Figure \ref{fig:csrate}.
Similar results can be obtained on the Brain, Cardiac and Artery images.

\subsection{Multi-contrast MRI}

\subsubsection{Multi-contrast MRI with wavelet joint sparsity}

To assist clinic diagnose, multiple MR images with different contrasts are often acquired simultaneously from the same anatomical cross section. For example, T1 and T2 weighted MR images
could distinguish fat and edema better, respectively. Different from the CS-MRI for individual MR imaging, multi-contrast
reconstruction for weighted MR images means the simultaneous reconstruction of multiple T1/T2-weighted MR images. Joint sparsity of the wavelet coefficients and JTV across different contrasts have been used in recent multi-contrast reconstruction methods \cite{Majumdar11MRI,huang2012fast}.

\begin{figure}[htbp]
\centering \vspace{-0.0cm}
 \includegraphics[scale=0.28]{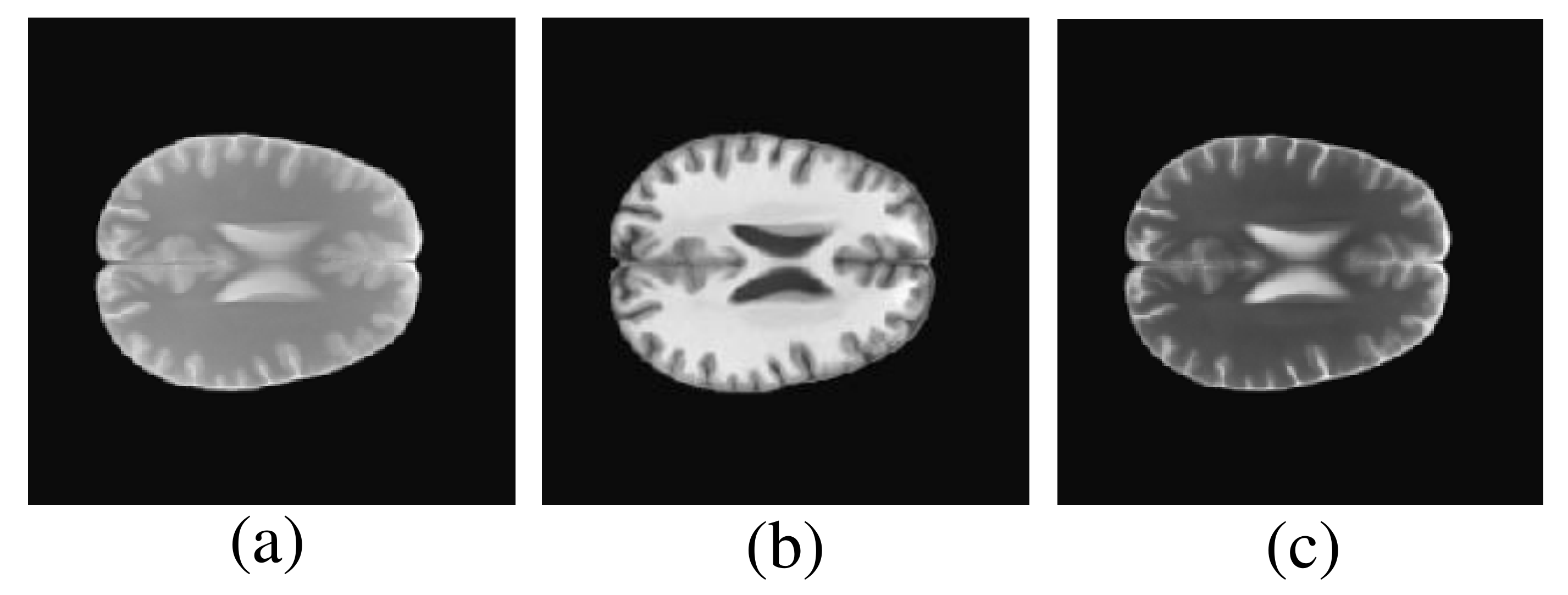}
\caption{The original images for multi-contrast MRI.}\label{fig:MTori}\vspace{-0.0cm}
\end{figure}

Here, the multi-contrast MR images are extracted
from the SRI24 Multi-Channel Brain Atlas Data \cite{Rohlfing10}. An example of the test images is shown in Fig. \ref{fig:MTori}.
We compare our method with the fastest multi-contrast MRI methods \cite{Majumdar11MRI,huang2012fast}, which use the algorithms SPGL1\_MMV \cite{Bergi08JSC} and FCSA to solve the corresponding problems, respectively. The experiment setup is the similar as in the previous experiments, except group setting is constructed for joint sparsity (non-overlapping) case.
FCSA\_MT and FIRLS\_MT denotes the algorithm in \cite{huang2012fast} and the proposed method in this setting.

\begin{figure}[htbp]
\centering \vspace{-0.0cm}
    \subfigure[]{\label{fig:2D}
        \includegraphics[scale=0.27]{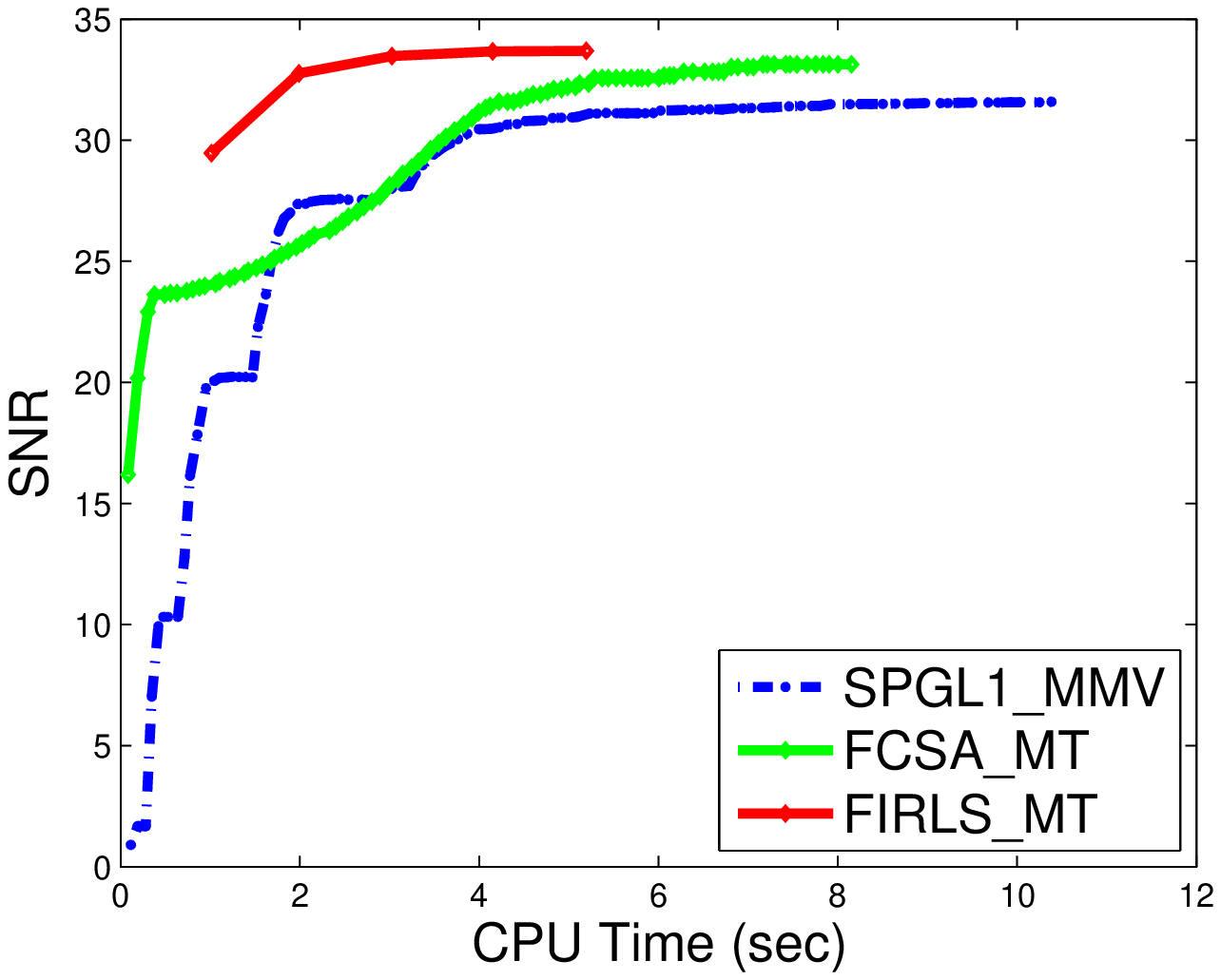}}
    \subfigure[]{\label{fig:2D}
        \includegraphics[scale=0.27]{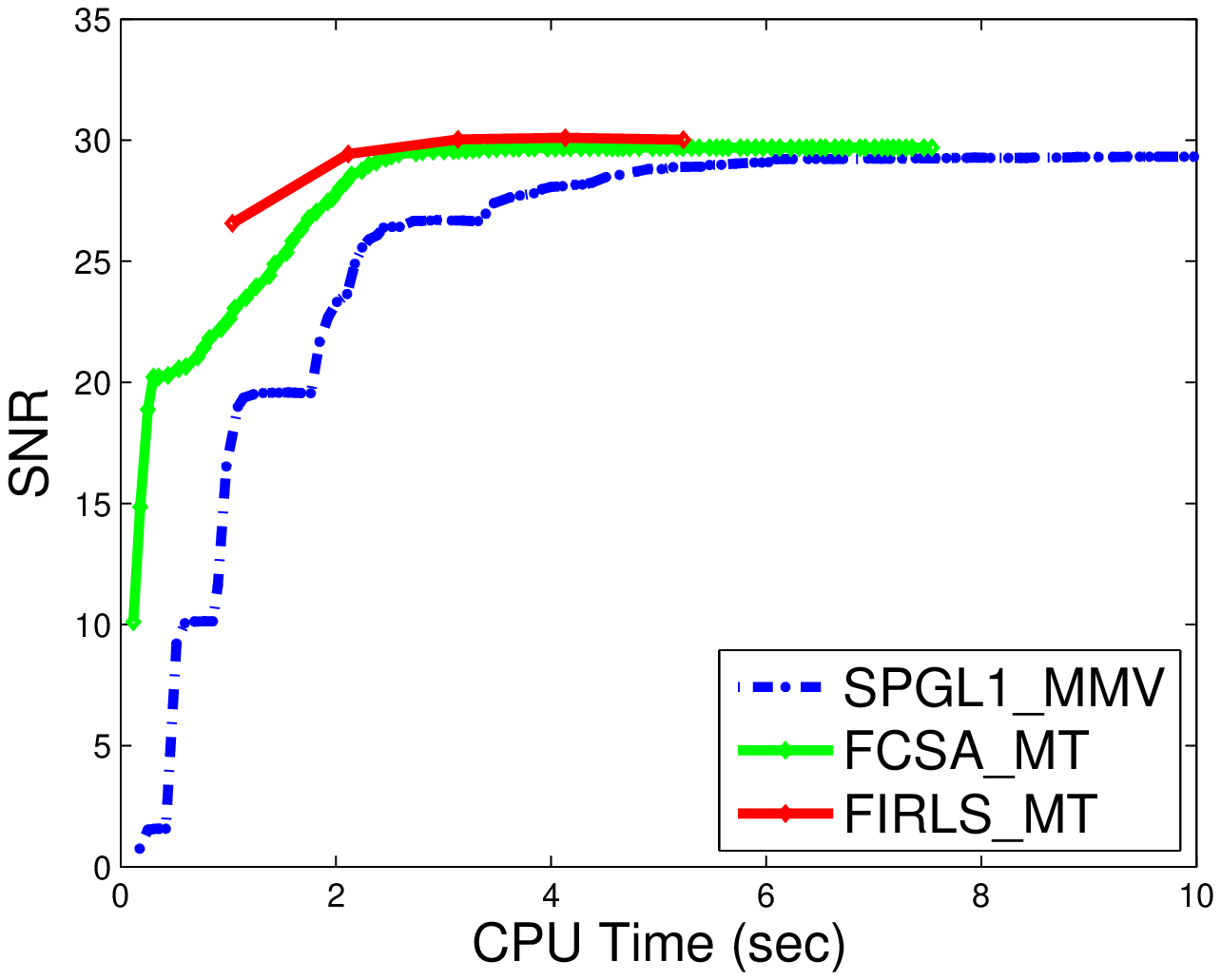}}
\caption{(a) Performance comparison for multi-contrast MRI with $25\%$ sampling. The average time costs of SPGL1\_MMV, FCSA\_MT, and the proposed method are 10.38 seconds, 8.15 seconds, 5.19 seconds. Their average SNRs are 31.58, 33.12 and 33.69. (b) Performance comparison for multi-contrast MRI with $20\%$ sampling. Their average time costs are 9.98 seconds, 7.54 seconds, 5.23 seconds. Their average SNRs are 29.31, 29.69 and 30.01.}\label{fig:MT}\vspace{-0.0cm}
\end{figure}

Fig. \ref{fig:MT} shows the performance comparisons among SPGL1\_MMV \cite{Bergi08JSC}, FCSA\_MT \cite{huang2012fast} and FIRLS\_MT on the example images shown in Figure \ref{fig:MTori}. Each algorithm runs 100 iterations in total. 
After convergence, three algorithms achieve similar accuracy for $20\%$ sampling and SPGL1 is only slightly worse than others for $25\%$ sampling.
From the curves, our method always ourperforms SPGL1\_MMV and FCSA\_MT, i.e., higher accuracy for the same reconstruction time.



\vspace{-0.0cm}
\begin{figure}[htbp]
\centering \vspace{-0.0cm}
    \subfigure[]{\label{fig:2D}
        \includegraphics[scale=0.27]{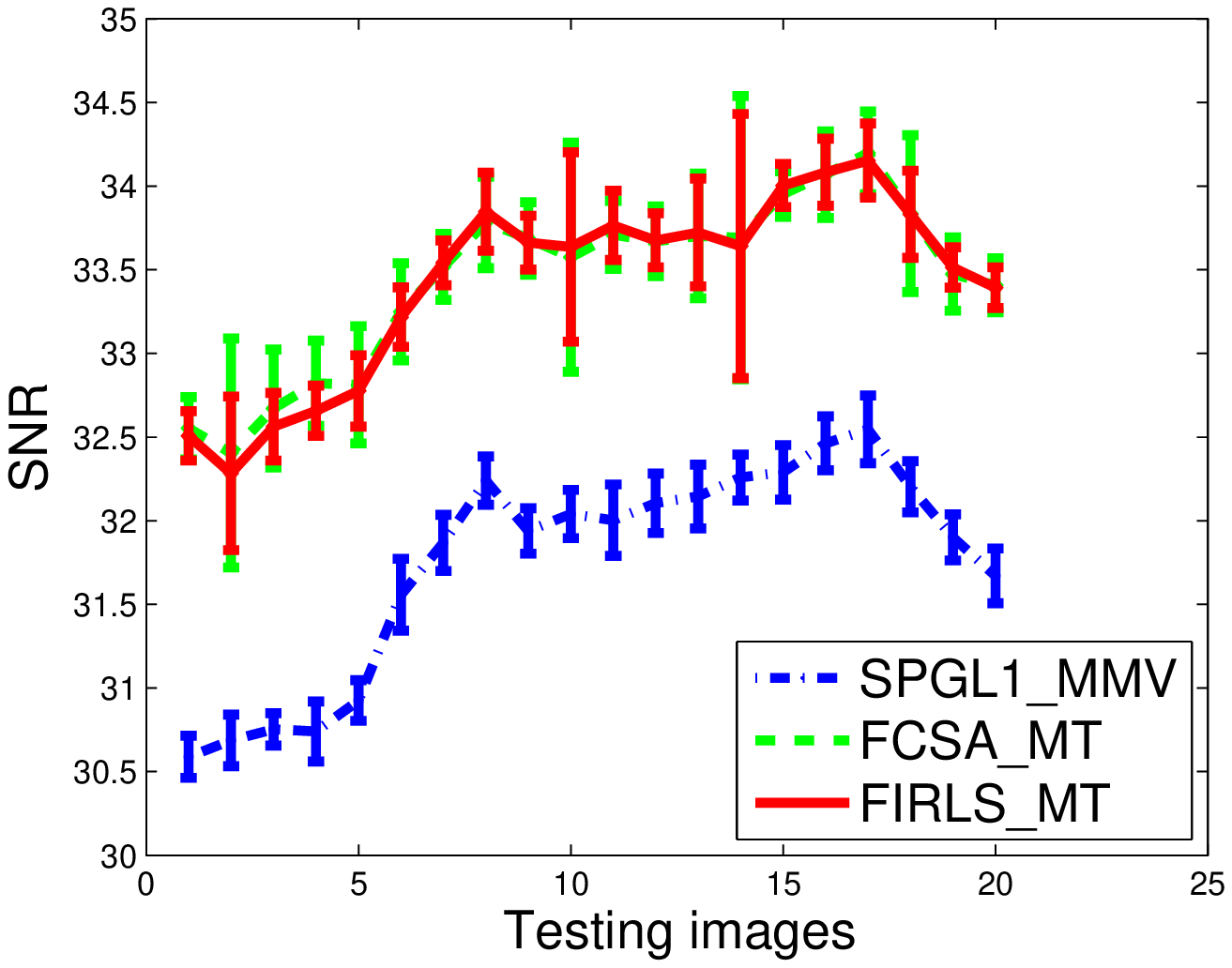}}
    \subfigure[]{\label{fig:2D}
        \includegraphics[scale=0.27]{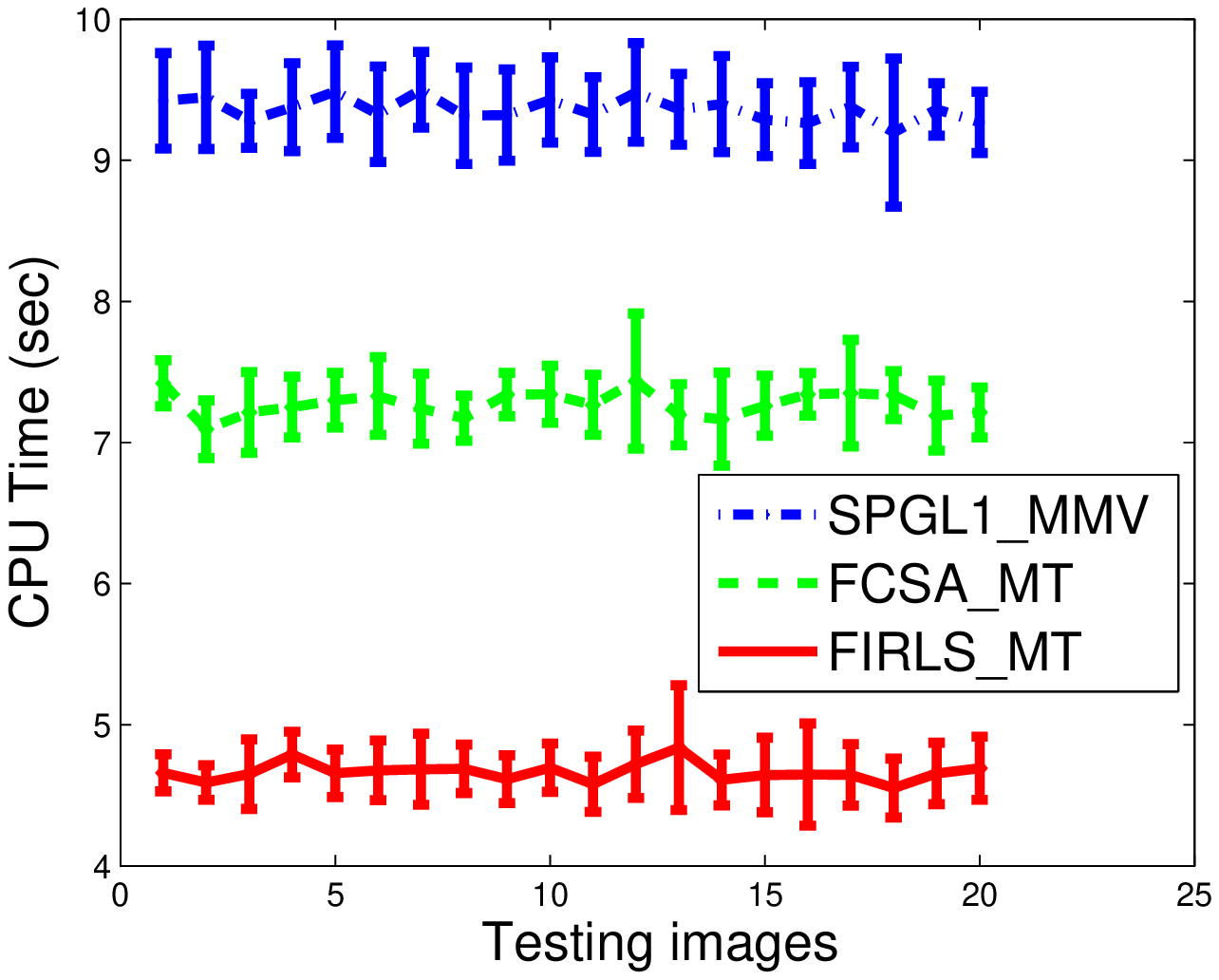}}
\caption{Performance comparison on 60 images from SRI24 dataset with $25\%$ sampling. (a) SNR comparison. (b) CPU time comparison. The average convergence time for SPGL1, FCSA\_MT and the proposed FIRLS\_MT is 9.3 seconds, 7.2 seconds, 4.6 seconds, respectively.}\label{fig:MTall}\vspace{-0.0cm}
\end{figure}

 \vspace{-0.0cm}
To quantitatively compare the convergence speed of these three methods, we conduct experiments on 20 set images (i.e. total 60 images) that are from SRI24. Different from the tree-based CS-MRI, each algorithm for non-overlapping group sparsity converges much faster. 
To reduce randomness, all algorithms run 100 times and the reconstruction results are shown in Fig. \ref{fig:MTall}. With $25\%$ sampling, the accuracy of our method is almost the same as FCSA\_MT, and always better than SPGL1. In the process to achieve the convergence, our method is consistently faster than the other two algorithms. These results demonstrate the efficiency of proposed method.

\vspace{-0.0cm}
 \vspace{-0.0cm}

\subsection{Discussion}


The first and second experiments validate the fast convergence speed of our method due to the proposed preconditioner. The advantages over the state-of-the-arts are further validated on practical application CS-MRI with four sparsity patterns: overlapping groups with tree sparsity, non-overlapping groups with joint sparsity, and TV. Although results on these problems are promising, some difference can be found. The non-overlapping group sparsity problem is often easier to solve. For example, the subproblem in FISTA has the closed form solution for joint sparsity but not for overlapping group sparsity. However, our method has similar difficulty for non-overlapping and overlapping group sparsity. That is why our method outperforms the fastest methods on joint sparsity reconstruction, and significantly outperforms those for tree-sparsity reconstruction, TV reconstruction. We do not compare the performance between the wavelet transform and and TV, since their performances are data-dependent. In this work, we only focus on fast minimization of the given functions. 

The superior performance of the proposed preconditioner also attributes to the structure of the system matrix $S$, which is often diagonally dominant in reconstruction problems (e.g. $A$ is random projection or partial Fourier transform). It can be applied to other applications where $S$ is not diagonally dominant (e.g. image blurring), and will be still more accurate than Jacobi preconditioner as it keeps more non-diagonal information.

\section{Conclusion}

%
%
%
We have proposed a novel method for analysis-based sparsity reconstruction, which includes structured sparsity with an orthogonal basis and total variation.
It is of the IRLS type and preserves the \emph{fast convergence rate}. The subproblem in our scheme is accelerated by the PCG method with a new pseudo-diagonal preconditioner. Due to the \emph{high accuracy} and \emph{efficiency} of this preconditioner, the subproblem can be solved in very low cost, even when it contains transforming operations. Extensive experimental results have demonstrated the \emph{flexibility}, \emph{effectiveness} and \emph{efficiency} of this method on CS-MRI.

\ifCLASSOPTIONcaptionsoff
  \newpage
\fi



%
{
\bibliographystyle{IEEEtran}
\bibliography{IEEEabrv,egbib}
}

\begin{IEEEbiography}[{\includegraphics[width=1in,height=1.25in,clip,keepaspectratio]{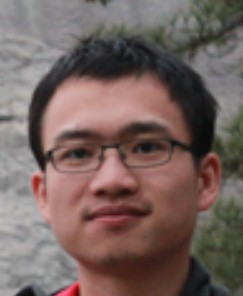}}]{Chen Chen}
received the BE degree and MS degree both from Huazhong University of Science and Technology, Wuhan, China, in 2008 and 2011, respectively.
He has been a PhD student in the Department of Computeter Science and Engineering at the University of Texas at Arlington since 2012.
His major research interests include image processing, medical imaging, computer vision and machine learning. He is a student member of the IEEE.
\end{IEEEbiography}

\begin{IEEEbiography}[{\includegraphics[width=1in,height=1.25in,clip,keepaspectratio]{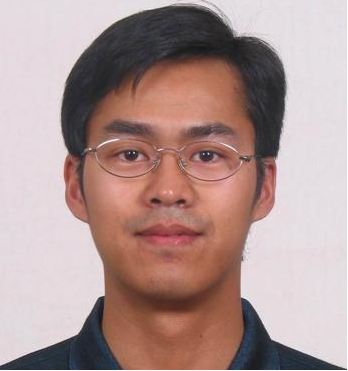}}]{Junzhou Huang}
received the BE degree from
Huazhong University of Science and Technology, Wuhan, China, in 1996, the MS degree
from the Institute of Automation, Chinese Academy of Sciences, Beijing, China, in 2003, and
the PhD degree from Rutgers University, New
Brunswick, New Jersey, in 2011. He is an
assistant professor in the Computer Science
and Engineering Department at the University of
Texas at Arlington. His research interests
include biomedical imaging, machine learning and computer vision,
with focus on the development of sparse modeling, imaging, and
learning for large scale inverse problems. He is a member of the IEEE.
\end{IEEEbiography}

\begin{IEEEbiography}[{\includegraphics[width=1in,height=1.25in,clip,keepaspectratio]{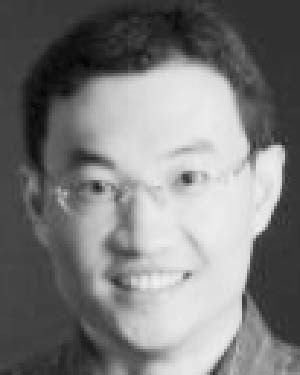}}]{Lei He}
received the PhD degree in electrical
engineering from the University of Cincinnati in
2003. Currently, he is a digital imaging scientist
at the Library of Congress, working on image
quality analysis and imaging devices evaluation
and calibration. From 2009 to 2011, he visited
the National Institutes of Health (NIH), working
on histology image analysis for cervical cancer
diagnosis. Before his visiting to the NIH, he was
an associate professor in the Department of
Computer Science and Information Technology at Armstrong Atlantic
State University in Savannah, Georgia. His research interests include
image processing and computer vision, and pattern recognition.
\end{IEEEbiography}

\begin{IEEEbiography}[{\includegraphics[width=1in,height=1.25in,clip,keepaspectratio]{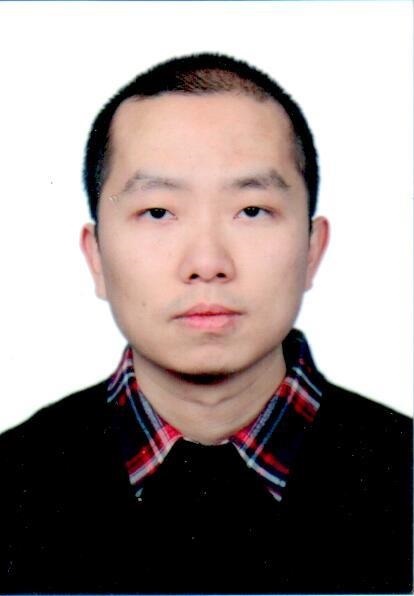}}]{Hongsheng Li}
received the bachelor¡¯s degree in automation from East China University of Science
and Technology, and the master¡¯s and doctorate degrees in computer science from Lehigh University,
Pennsylvania, USA in 2006, 2010 and 2012, respectively. He is an associate professor in the School of
Electronic Engineering at University of Electronic Science and Technology of China. His research interests
include computer vision, medical image analysis and machine learning.
\end{IEEEbiography}



\enlargethispage{-5in}

\end{document}